\documentclass[11pt]{article}

\usepackage[final]{acl}

\usepackage{times}
\usepackage{latexsym}

\usepackage[T1]{fontenc}

\usepackage[utf8]{inputenc}

\usepackage{microtype}

\usepackage{inconsolata}

\usepackage{graphicx}

\usepackage{amsmath,amsfonts,bm}

\def\eqref#1{equation~\ref{#1}}

\def\1{\bm{1}}

\DeclareMathAlphabet{\mathsfit}{\encodingdefault}{\sfdefault}{m}{sl}
\SetMathAlphabet{\mathsfit}{bold}{\encodingdefault}{\sfdefault}{bx}{n}

\usepackage{booktabs} 
\usepackage{multirow}
\usepackage{arydshln}
\usepackage{graphicx}
\usepackage{enumitem}
\usepackage{url}
\usepackage{colortbl}
\usepackage{xcolor}
\usepackage{natbib}
\usepackage{tcolorbox}
\tcbuselibrary{most}
\usepackage{tikz}
\usepackage{footnote}
\makesavenoteenv{table}
\usepackage{tikz}
\usepackage{tabularx}
\usepackage{xcolor}
\usepackage{amsmath}
\usepackage[utf8]{inputenc}
\usepackage{amssymb}
\newcommand{\dashedrect}{%
    \tikz[baseline=0.2ex]{%
        \draw[black, dashed, line width=0.8pt] (0,0) rectangle (5mm, 2.5mm);%
        \path[use as bounding box] (-0.8mm, 0) rectangle (5.8mm, 2.5mm);%
    }%
}
\newcommand{\solidrect}{%
    \tikz[baseline=0.2ex]{%
        \draw[black, solid, line width=0.8pt] (0,0) rectangle (5mm, 2.5mm);%
        \path[use as bounding box] (-0.8mm, 0) rectangle (5.8mm, 2.5mm);%
    }%
}
\usepackage{makecell}

\newcommand{\think}[1]{\textcolor{blue}{\texttt{<think>}} #1 \textcolor{blue}{\texttt{</think>}}}
\newcommand{\asking}[1]{\textcolor{cyan}{\texttt{<asking>}} #1 \textcolor{cyan}{\texttt{</asking>}}}
\newcommand{\response}[1]{\textcolor{brown}{\texttt{<response>}} #1 \textcolor{brown}{\texttt{</response>}}}

\title{Reasoning While Asking: Transforming Reasoning Large Language Models from Passive Solvers to Proactive Inquirers}

\author{
 \textbf{Xin Chen\textsuperscript{1,2,3,*}}, ~~
 \textbf{Feng Jiang\textsuperscript{2,}\Thanks{Equal contribution}}, ~~
 \textbf{Yiqian Zhang\textsuperscript{3}}, ~~
 \textbf{Hardy Chen\textsuperscript{4}}, ~~
 \textbf{Shuo Yan\textsuperscript{5}}, ~~
 \\
 \textbf{Wenya Xie\textsuperscript{6}}, ~~
 \textbf{Min Yang\textsuperscript{2,3,†}}, ~~
 \textbf{Shujian Huang\textsuperscript{1,}\Thanks{Corresponding author.}}
 \vspace{2mm}
\\ 
 \textsuperscript{1}National Key Laboratory for Novel Software Technology, Nanjing University, \\
 \textsuperscript{2}Artificial Intelligence Research Institute, Shenzhen University of Advanced Technology, \\
 \textsuperscript{3}Shenzhen Institutes of Advanced Technology, Chinese Academy of Sciences, \\
 \textsuperscript{4}University of California, Santa Cruz,
 \textsuperscript{5}University of Texas, Dallas, 
 \textsuperscript{6}University of Minnesota 
 \\
\texttt{x.chen@smail.nju.edu.cn, jiangfeng@suat-sz.edu.cn} \\
\texttt{min.yang@siat.ac.cn, huangsj@nju.edu.cn}
}
\begin{document}
\maketitle
\begin{abstract}
Reasoning-oriented Large Language Models (LLMs) have achieved remarkable progress with Chain-of-Thought (CoT) prompting, yet they remain fundamentally limited by a \emph{blind self-thinking} paradigm: performing extensive internal reasoning even when critical information is missing or ambiguous. 
We propose Proactive Interactive Reasoning (PIR), a new reasoning paradigm that transforms LLMs from passive solvers into proactive inquirers that interleave reasoning with clarification. 
Unlike existing search- or tool-based frameworks that primarily address knowledge uncertainty by querying external environments, PIR targets premise- and intent-level uncertainty through direct interaction with the user. PIR is implemented via two core components: 
(1) an uncertainty-aware supervised fine-tuning procedure that equips models with interactive reasoning capability, and (2) a user-simulator-based policy optimization framework driven by a composite reward that aligns model behavior with user intent. Extensive experiments on mathematical reasoning, code generation, and document editing demonstrate that PIR consistently outperforms strong baselines, achieving up to 32.70\% higher accuracy, 22.90\% higher pass rate, and 41.36 BLEU improvement, while reducing nearly half of the reasoning computation and unnecessary interaction turns.
Further reliability evaluations on factual knowledge, question answering, and missing-premise scenarios confirm the strong generalization and robustness of PIR. 
\footnote{Model and code are publicly available at: \href{https://github.com/SUAT-AIRI/Proactive-Interactive-R1}{https://github.com/SUAT-AIRI/Proactive-Interactive-R1}}

\end{abstract}

\section{Introduction}

\begin{figure*}[!t]
    \centering
    \includegraphics[width=0.95\textwidth, trim=0 0 0 0]{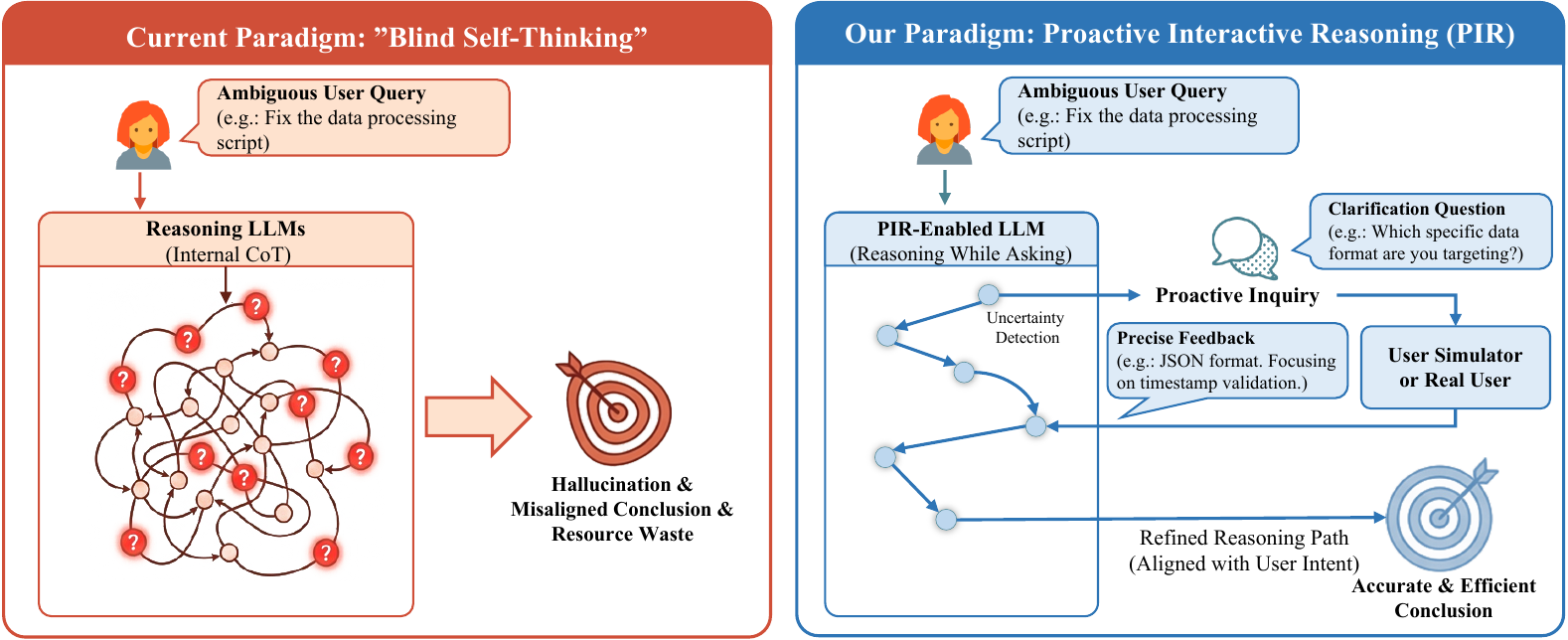}
    \caption{\textbf{The Proactive Interactive Reasoning (PIR) Paradigm.} The schematic contrasts inefficient "blind self-thinking" on ambiguous queries with the PIR approach. PIR utilizes uncertainty detection and a two-phase optimization mechanism to enable proactive clarification with a user simulator, aligning reasoning chains with user intent to achieve accurate problem-solving that is \textbf{efficient}, \textbf{robust}, and \textbf{minimal compute costs.}}
    \label{fig:demo}
    \vspace{-1.2em}
\end{figure*}

The emergence of reasoning-oriented LLMs such as GPT-o1~\citep{openai2024openaio1card} and DeepSeek-R1~\citep{guo2025deepseek} marks a fundamental paradigm shift toward self-thinking models, which internally deliberate through multi-step CoT reasoning. This paradigm has substantially boosted performance on complex tasks, and recent studies further suggest that the quality of explicit reasoning traces is positively correlated with final accuracy~\citep{DBLP:journals/corr/abs-2504-13367}.

Despite these advances, current reasoning LLMs exhibit a critical limitation, which we term blind self-thinking: they often fail to recognize when their reasoning is underinformed or based on ambiguous user instructions~\citep{DBLP:journals/corr/abs-2401-08329}, as shown in the left part of \autoref{fig:demo}. When users provide incomplete or ambiguous prompts, a common occurrence in real-world interactions, models tend to continue lengthy reasoning regardless, leading to overthinking~\citep{DBLP:journals/corr/abs-2502-08235}, hallucinations~\citep{DBLP:conf/emnlp/YuanCJ0Z0024}, and misaligned conclusions~\citep{fan2025missing,chen2025do}. Users are then forced to perform iterative post-hoc corrections, which significantly degrade interaction efficiency and user experience~\citep{DBLP:journals/corr/abs-2401-08329,DBLP:conf/iui/KimL0PK24}. Although recent multi-turn interactive LLMs~\citep{collabllm2025} acknowledge this by attempting to reactively follow user feedback, they remain passive solvers and fail to actively reduce unnecessary conversational turns.

To bridge the gap between ambiguous user queries and precise reasoning execution, we introduce PIR, a new paradigm that transforms reasoning LLMs from passive solvers into proactive inquirers, offering a principled approach for aligning reasoning with user intent under ambiguity, as shown in the right of \autoref{fig:demo}. Instead of relying on user-initiated corrections, PIR explicitly trains models to detect missing premises and autonomously initiate clarification during the reasoning process.

Our approach consists of two clearly separated stages. First, we propose an uncertainty-aware mechanism for constructing an interactive reasoning dataset to activate the model's interactive capability. Specifically, the mechanism detects critical decision points where the model’s confidence dips. At these junctures, we convert monologic reasoning traces into a think-and-ask format by injecting clarification questions and user replies simulated by instruction-following LLMs. Then, to further align the model’s reasoning behavior with user intent, we introduce a novel Group Relative Policy Optimization framework equipped with a dynamic user simulator (US-GRPO) and a principled composition of extrinsic (task success) and intrinsic (helpfulness-efficiency) rewards. This optimization explicitly encourages the model to prioritize intent resolution over ungrounded self-thinking, leading to significant improvements in both correctness and interaction efficiency.

Extensive experiments on mathematical reasoning, code generation, and document editing show that PIR effectively avoids invalid reasoning trajectories, reducing computation by an average of approximately 2k tokens per task and cutting unnecessary interaction turns by half. Meanwhile, PIR achieves superior performance with improvements of 9.8\% in Accuracy, 3.2\% in Pass Rate, and 13.36 in BLEU, respectively. Additionally, PIR also demonstrates strong generalization on non-interactive benchmarks, including factual knowledge, question answering, and missing premise tests, suggesting that proactive interactive reasoning is beneficial beyond interactive settings.

Our contributions are summarized as follows:
\begin{itemize}[label=\textbullet, topsep=1pt, itemsep=1pt, left=1pt]
\item We identify and formalize the \textbf{blind self-thinking} problem in current reasoning LLMs and propose the \textbf{PIR} framework to enable proactive clarification.
\item We develop a reinforcement learning method, \textbf{US-GRPO}, incorporating a dynamic user simulator and both extrinsic and intrinsic rewards to align reasoning with intent resolution and optimize interaction efficiency.
\item Through extensive experiments, we show that PIR achieves strong performance on interactive benchmarks and demonstrates robust generalization across non-interactive reasoning settings, providing a more efficient and user-aligned paradigm for next-generation reasoning models.
\end{itemize}
\section{Related Work}
\begin{figure*}[!ht]
    \centering
    \includegraphics[width=0.95\textwidth, trim=0 0 0 0]{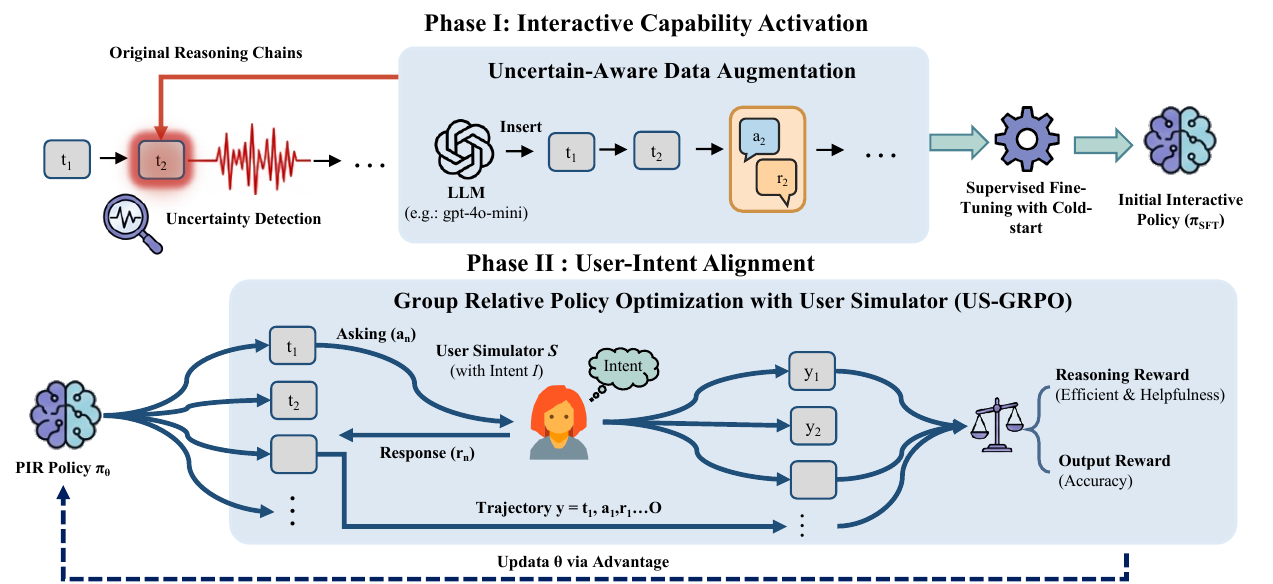}
    \caption{\textbf{Overview of the PIR Framework.} The framework operates in two phases to transition LLMs from passive solvers to proactive inquirers.}
    \label{fig:methodology}
    \vspace{-1.2em}
\end{figure*}
\subsection{Reasoning Large Language Model}
Recent reasoning LLMs, such as GPT-o1~\citep{openai2024openaio1card} and DeepSeek-R1~\citep{guo2025deepseek}, have shown impressive performance on complex problem-solving tasks by generating explicit CoT before producing final answers. Despite these advances, these models frequently exhibit pathological reasoning behaviors: they may overthink~\citep{fan2025missing,chen2025do}, even hallucinate during intermediate reasoning~\citep{DBLP:conf/emnlp/YuanCJ0Z0024}, or produce misaligned conclusions that waste substantial computation~\citep{DBLP:journals/corr/abs-2511-04108}. 

To alleviate these issues, a growing line of research attempts to augment self-thinking models with external grounding mechanisms. Retrieval-augmented approaches such as Search-R1~\citep{jin2025searchr} and related variants~\citep{song2025r1searcherincentivizingsearchcapability,luo2025graphr1} incorporate retrieval into the reasoning process to reduce factual mistakes by supplying verified external evidence. Other works~\citep{qian2025toolrl,zhang2025toolr1sampleefficientreinforcementlearning} empower reasoning with external tools, e.g., code interpreters or APIs, to validate intermediate steps or execute computations. While these approaches effectively mitigate factual hallucinations, they tend to treat errors primarily as knowledge deficiencies rather than intent deficiencies.  As a result, they remain insufficient when the primary source of failure is misunderstanding the user's intent.

\subsection{Interactive Large Language Model}
A complementary research direction focuses on enhancing LLM interactivity to improve user–model alignment. Early methods~\citep{kim-etal-2023-tree,DBLP:conf/emnlp/FaltingsGBPC0GD23,DBLP:journals/corr/abs-2405-15784,DBLP:conf/www/ZhaoD24} explored prompting strategies that guide LLMs to ask clarification questions. However, these methods often depend on fixed interaction templates, limiting flexibility and generalization across diverse tasks.

Recent studies have investigated more principled training approaches. CLAM~\citep{kuhn2023clamselectiveclarificationambiguous} demonstrates substantial gains by using synthetic ambiguous queries generated via controlled perturbations. Other works~\citep{DBLP:journals/corr/abs-2311-02737,DBLP:conf/www/ZhaoD24} employ reinforcement learning to teach LLMs when to issue clarification requests. Yet, follow-up analyses~\citep{zhang-etal-2024-clamber} reveal that existing systems frequently struggle to determine what information is missing and which question would effectively reduce ambiguity, suggesting inadequacies in their reasoning processes. CollabLLM~\citep{collabllm2025} reframes LLMs as active collaborators rather than passive solvers. While effective, this line of work still relies largely on surface-level behavioral patterns. Crucially, these models do not explicitly integrate the clarification step into the internal reasoning chain, leaving the decision-making behind “when and why to ask” fundamentally opaque.

\section{Proactive Interactive Reasoning Framework}

To address the limitations of blind self-thinking, we propose the PIR framework, which equips reasoning LLMs with the ability to proactively query users and dynamically integrate feedback during reasoning. As shown in Figure~\ref{fig:methodology}, PIR consists of two phases:  
(1) \textbf{Interactive Capability Activation}, where the model learns \textit{when and how} to ask; and  
(2) \textbf{User-Intent Alignment}, where the model refines its reasoning based on user feedback.

\subsection{Phase I: Interactive Capability Activation}
A core challenge in activating interactive capability is identifying when the model should initiate clarification. In real-world reasoning, the need for clarification often emerges when the model becomes uncertain about its intermediate decisions, yet continues to reason without sufficient information. Therefore, we argue that an effective training signal for interactive reasoning should be closely tied to the model’s internal uncertainty during the reasoning process.

\subsubsection{Uncertainty-Aware Interactive Data Augmentation}
Inspired by prior work~\citep{yang2025understandingahamomentsexternal} showing that reasoning LLMs exhibit spikes in uncertainty around ``aha moments'', indicating points where the model is unsure, we leverage this observation to convert uncertainty into a \textit{direct training signal} for interactive reasoning.

Given an input $\mathbf{x}$, we obtain the initial reasoning traces using a frozen teacher model and segment it into a sequence of reasoning steps $\mathbf{s}=\{s_1,\dots,s_n\}$, where each step corresponds to a semantically coherent sentence-level unit. For each step $s_j$, we compute its Predictive Entropy (PE)~\citep{DBLP:journals/corr/abs-2207-05221} as
\begin{equation}
\vspace{-2mm}
\small
\mathrm{Uncertainty}(s_j \mid \mathbf{x}) = -\frac{1}{|s_j|} \sum_{i=1}^{|s_j|} \log p(z_i \mid z_{<i}, s_{<j}, \mathbf{x})
\end{equation}
where $z_i$ denotes the $i$-th token within step $s_j$. Steps whose $\mathrm{Uncertainty}$ values rank in the top-$k\%$ are selected as candidate clarification points. At each selected clarification point, we employ a strong instruction-following model (e.g., GPT-4o-mini) to insert an explicit clarification question and simulate a corresponding user response. This process transforms an originally linear CoT into an interleaved \textit{think–ask–respond} trajectory:
\begin{equation}
\small
\vspace{-2mm}
y = \{t_1, (a_1, r_1), \dots, t_m, (a_m, r_m), O\}
\end{equation}
where $t_i$ denotes a reasoning step, $(a_i, r_i)$ is the inserted clarification–response pair, and $O$ is the final answer.  
Through this construction, the model is explicitly exposed to the causal pattern
\textit{uncertainty} $\rightarrow$ \textit{clarification} $\rightarrow$ \textit{improved reasoning},
which serves as a direct training signal for learning proactive interactive behavior.

\subsubsection{Supervised Fine-Tuning for Cold-Start}
To endow the model with the interactive capability for generating clarifying questions during thinking, we perform supervised fine-tuning (SFT) on the augmented trajectories. Given input $x$ and its interaction sequence $y = y_{1:L}$, the training objective is the standard autoregressive loss:
\begin{equation}
\vspace{-2mm}
\small
\mathcal{L}_{\mathrm{SFT}}
=
-\sum_{j=1}^{L} 
\log p_\theta(y_j \mid x, y_{<j})
\end{equation}

The optimization spans the entire sequence $y$, including reasoning steps $(t_i)$, proactive questions $(a_i)$, simulated responses $(r_i)$, and the final answer $(O)$, allowing the model to learn smooth transitions between internal reasoning, inquiry initiation, and feedback incorporation.

\subsection{Phase II: User-Intent Alignment}
While Phase I equips the model with the structural capability to initiate interaction, it does not explicitly optimize whether a clarification is necessary nor how to ask efficiently, as reflected in \autoref{tab:main_result_transposed}. Moreover, optimizing such behavior directly with real users is impractical due to high cost, limited availability, and uncontrolled noise. Therefore, we construct US-GRPO, a controllable interactive environment featuring a user simulator, which enables the systematic optimization of interactive reasoning behaviors with a designed composite reward, thereby promoting accurate, efficient, and intent-aligned problem solving.

\subsubsection{Group Relative Policy Optimization with User Simulator}
\paragraph{User Simulator Construction.}
We first build a dynamic interactive environment using an instruction-following LLM prompted as a user simulator $\mathcal{S}$, conditioned on a specific user intent $\mathcal{I}$ (defined in Appendix~\ref{app:prompt_template}). Within this environment, the interactive reasoning policy $\pi_\theta$ generates reasoning steps and clarification questions, while the simulator produces responses that are consistent with the underlying intent.

Given an input $x$ and intent $\mathcal{I}$, the joint interactive trajectory $y$ is generated as
\begin{equation}
\small
\label{eq:user_simulator}
\resizebox{0.9\linewidth}{!}{$
y \sim p(y \mid x, \mathcal{I})
= \prod_{n=1}^{N} \pi_\theta(t_n, a_n \mid h_n) \cdot \mathcal{S}(r_n \mid a_n, \mathcal{I})
$}
\end{equation}
where $h_n$ denotes the interaction history at turn $n$, $t_n$ is the $n$-th reasoning segment, $a_n$ is the clarifying question (if any), and $r_n$ is the simulator's response. This formulation enables the model to explore diverse questioning strategies in a controllable, responsive environment.

\paragraph{Optimization with GRPO.}

To optimize the objective, we adopt GRPO~\citep{DBLP:journals/corr/abs-2402-03300}, an efficient reinforcement learning (RL) algorithm that avoids training a separate value-function critic.

For each query $x$, GRPO samples a group of $G$ trajectories $\{y_1, \dots, y_G\}$ from the old policy $\pi_{\mathrm{old}}$. The policy is updated by maximizing
\begin{equation}
\vspace{-2mm}
\small
\resizebox{0.9\linewidth}{!}{$
\begin{aligned}
\mathcal{J}_{\mathrm{GRPO}}(\theta)
= \mathbb{E}_{x \sim \mathcal{D},\, y_i \sim \pi_{\mathrm{old}}} 
\Bigg[ \frac{1}{G} \sum_{i=1}^{G} 
\Big( \prod_{n=1}^{N_i} 
\frac{\pi_\theta(t_{n}^{(i)}, a_{n}^{(i)} \mid h_{n}^{(i)})}
{\pi_{\mathrm{old}}(t_{n}^{(i)}, a_{n}^{(i)} \mid h_{n}^{(i)})}
\Big) \hat{A}_i \\
\qquad\qquad\quad
- \beta\, \mathbb{D}_{\mathrm{KL}}(\pi_\theta \,\|\, \pi_{\mathrm{ref}}) \Bigg]
\end{aligned}
$}
\vspace{-2mm}
\end{equation}
where $\hat{A}_i$ is the group-relative advantage, and $\pi_{\mathrm{ref}}$ is a reference policy that regularizes updates via KL divergence. During optimization, gradients are only applied to the policy outputs $(t_n, a_n)$, while $r_n$ is fully masked out from the policy gradient.

\subsubsection{Composite Reward Modeling}
Our optimization objective is to encourage the model to produce correct final answers while minimizing unnecessary interaction and ensuring high-quality clarification behavior. To this end, we define a composite reward function that evaluates both the final output and the interaction process. Let $y$ be a generated trajectory consisting of a reasoning path $r$ (including interaction turns) and a final output $o$, with ground-truth answer $g$. The total reward is defined as: 
\begin{equation}
\vspace{-2mm}
\small
R(y) = R_{\mathrm{output}}(o, g) + R_{\mathrm{reason}}(r, o, g)
\end{equation}
\paragraph{Output Reward.}
The output reward $R_{\mathrm{output}}$ ensures basic task correctness. It assigns a baseline score $S_{\text{base}}$ only when the final answer matches the ground truth:
\begin{equation}
\vspace{-2mm}
\small
R_{\mathrm{output}}(o, g) = S_{\text{base}} \cdot \mathbb{I}(o = g)
\end{equation}
where $\mathbb{I}(\cdot)$ is the indicator function.
\paragraph{Reasoning Reward.}
The reasoning reward $R_{\mathrm{reason}}$ evaluates the quality of the interaction process and is deliberately activated \emph{only when the answer is correct}, so as not to reward hallucinated or misleading reasoning. For a reasoning path $r$ that involves $n$ clarification turns, we define
\begin{equation}
\small
\resizebox{0.9\linewidth}{!}{$
R_{\mathrm{reason}}(r, o, g)
= \mathbb{I}(o = g) \cdot \mathbb{I}_{\text{ask}}
\Big[ S_{\text{base}} \cdot E(r) \cdot H_{\text{LLM}}(r) \Big]
$}
\end{equation}
where $E(r)$ measures interaction efficiency and $H_{\text{LLM}}(r) \in [0,1]$ evaluates the helpfulness of clarification turns, derived from an evaluation by a strong LLM acting as a judge based on the prompt provided in Appendix~\ref{app:prompt_template}. This design naturally shapes clarification behavior. When information is missing, direct answers fail ($\mathbb{I}(o=g)=0$), so the model learns to clarify. When queries are clear, the model is rewarded for answering directly. We formulate $E(r)$ as
\begin{equation}
\small
E(r) = \frac{N_{\text{max}} - n}{N_{\text{max}} - 1}
\end{equation}
where $n$ is the number of clarification turns and $N_{\text{max}}$ is the maximum allowed number of turns.

\section{Simulated Experiments}

\subsection{Experimental Settings}

\subsubsection{Datasets}
To enable proactive interaction at cold start, we construct a new \textbf{Reasoning-while-asking} SFT dataset. It is built upon open-ended questions from~\cite{DBLP:journals/corr/abs-2304-07987} and is specifically designed to teach models \emph{when and how to ask clarification questions during reasoning}. Concretely, we employ DeepSeek-R1~\cite{guo2025deepseek} as a frozen reference model to generate initial reasoning trajectories, as described in Appendix~\ref{app:prompt_template}, forming interactive \emph{reasoning-while-asking} training samples. Our initial analysis in Appendix~\ref{app:uncertainty_anlysis} demonstrates that the uncertainty-aware mechanism effectively enables the model to pinpoint high-uncertainty clarification points and master the interactive format. 

To further align model behavior with user intent, we adopt three publicly available multi-turn task datasets from~\cite{collabllm2025}: \textbf{Math-Chat}, \textbf{BigCodeBench-Chat}, and \textbf{DocEdit-Chat}, which respectively represent multi-turn mathematical problem solving, coding assistance, and collaborative document editing—three of the most common real-world usage scenarios of LLMs. These datasets are used in the RL stage. We split each dataset into training and test sets for RL evaluation; detailed statistics and splits are reported in Appendix~\ref{app:detail_of_training_cofig}.

\begin{table*}[!ht]
\centering
\resizebox{0.98\linewidth}{!}{
\begin{tabular}{l| cccc | cccc | cccc}
\toprule
\multirow{4}{*}{\textbf{Method}} & 
\multicolumn{4}{c}{\textbf{MATH-Chat}} & 
\multicolumn{4}{c}{\textbf{BigCodeBench-Chat}} & 
\multicolumn{4}{c}{\textbf{DocEdit-Chat}} \\
\cmidrule(lr){2-5} \cmidrule(lr){6-9} \cmidrule(lr){10-13}
& ACC & Tokens(k) & TTR  & Help. & PR & Tokens(k) & TTR & Help. & BLEU & Tokens(k) & TTR & Help. \\
\midrule

\multicolumn{13}{c}{{\textbf{Multi-turn LLM}}} \\
Instruct Base~\cite{DBLP:journals/corr/abs-2409-12122} & 21.30 & 2.34 & 3.85 & 0.39 & 15.50 & 2.78 & 3.80 & 0.27  & 8.50 & 1.72 & 4.65 & 0.07  \\
\quad + Proactive Prompt & \underline{22.90} & 2.20 & 3.71 & \underline{0.41}  & \underline{19.70} & 2.84 & 3.99 & 0.21 & 7.10 & 2.14 & 4.11 & 0.05  \\
STaR-GATE~\cite{DBLP:journals/corr/abs-2403-19154} & 1.80 & 2.10 & 4.20 & 0.11 & 8.80 & \textbf{0.83} & 4.70 & 0.10 &  7.00 & \underline{0.81} & 4.90 & 0.06  \\
CollabLLM~\cite{collabllm2025} & 16.20 & \underline{1.99} & 3.65 & 0.39  & 10.20 & 2.87 & 4.82 & \underline{0.42} & 28.00 & 2.83 & 4.28 & \underline{0.48}  \\
\midrule

\multicolumn{13}{c}{{\textbf{Proactive Interactive Reasoning LLM}}} \\
Reasoning Base~\cite{guo2025deepseek} & 15.20 & 3.61 & \textbackslash & \textbackslash & 9.10 & 2.15 & \textbackslash & \textbackslash &  \underline{34.92} & 1.40 & \textbackslash & \textbackslash  \\
\quad + Active SFT (w/ self-play) & 15.30 & 3.26 & \textbackslash & \textbackslash  & 9.90 & 2.08 & \textbackslash & \textbackslash & 28.27 & 1.82  & \textbackslash & \textbackslash \\
\quad \quad + US-GRPO (w/ self-play) & 20.80 & 1.67 & \textbackslash & \textbackslash & 10.90 & \underline{1.07} & \textbackslash & \textbackslash & 41.05 & \textbf{0.80} & \textbackslash & \textbackslash   \\
\quad + Active SFT (w/ interactive) & 9.70 & 2.06 & \textbf{1.22} & 0.24 & 9.10 & 1.38 & \textbf{0.32}  & 0.22 & 24.75 & 2.41 & \underline{2.00}  & 0.36  \\
\quad \quad + US-GRPO (w/ interactive) & \textbf{32.70} & \textbf{1.70} & \underline{1.80}  & \textbf{0.44}  & \textbf{22.90} & 1.30 & 1.29  & \textbf{0.46} & \textbf{41.36} & 0.83 & \textbf{1.00}  & \textbf{0.66}  \\
\bottomrule
\end{tabular}
}
\caption{Evaluation results between the Multi-turn LLM and PIR LLM across Math-Chat, BigCodeBench-Chat, and DocEdit-Chat datasets. The \textbf{bold} indicates the best performance and the \underline{underline} indicates the second best.}
\label{tab:main_result_transposed}
\vspace{-1.2em}
\end{table*}
\subsubsection{Baselines}
We compare our method with two groups of models that reflect different modeling philosophies.

\paragraph{Multi-turn LLM.}
This group evaluates general conversational and interactive strategies on the non-reasoning LLM.
\begin{itemize}[label=\textbullet, topsep=1pt, itemsep=1pt, left=1pt]
    \item \textbf{Instruction-Tuned LLM}~\cite{DBLP:journals/corr/abs-2409-12122}. We use \texttt{Qwen2.5-Math-7B-Instruct} as a general-purpose multi-turn model without explicit reasoning or interaction enhancement.
    \item \textbf{Proactive Prompt}. Built upon the Instruction-Tuned LLM, this baseline introduces explicit clarification instructions (see Appendix~\ref{app:prompt_template}) to encourage proactive questioning.
    \item \textbf{STaR-GATE}~\cite{DBLP:journals/corr/abs-2403-19154}. An interactive method that trains LLMs to clarify when encountering ambiguity.
    \item \textbf{CollabLLM}~\cite{collabllm2025}. A collaborative framework that models long-term conversational contribution using multi-turn-aware RL.
\end{itemize}

\paragraph{PIR LLM.}
This group isolates the contributions of each component in our framework.
\begin{itemize}[label=\textbullet, topsep=1pt, itemsep=1pt, left=1pt]
    \item \textbf{Base Reasoning LLM}~\cite{guo2025deepseek}. We adopt \texttt{DeepSeek-R1-Distill-Qwen-7B} as the base reasoning model, evaluated in its zero-shot setting without explicit interactive tuning.
    \item \textbf{Active SFT}. The Base Reasoning LLM is further trained on our SFT dataset, enabling self-play and interactive completion.
    \item \textbf{US-GRPO}. The Active SFT model is subsequently optimized via US-GRPO with a dynamic user simulator. We additionally employ Decoupled Clip and Dynamic sAmpling Policy Optimization (DAPO)~\cite{yu2025dapo} to accelerate convergence and better align reasoning trajectories with user intent.
\end{itemize}

\subsubsection{Implementation Details}
\paragraph{Interactive Environment Construction.} Following the design of CollabLLM~\cite{collabllm2025}, we construct an interactive environment in which a general user simulator, \texttt{Llama-3.1-8B-Instruct}, is employed to generate initially ambiguous requests shown in the Appendix~\ref{app:prompt_template}, and the maximum number of interaction turns is set to 5 for all methods.

To ensure fair and rigorous comparison, different methods adopt different termination mechanisms.
For the Multi-turn LLM baselines, the user simulator is equipped with an explicit termination signal \texttt{[TERMINATE CHAT]}, which is triggered once the objective is achieved or further interaction yields negligible improvement. This design prevents evaluation bias caused by redundant generations.
In contrast, PIR LLM does not rely on any external termination signal, because it can learn to dynamically decide the appropriate number of interaction turns during training.

\paragraph{Evaluation Paradigm.}
We evaluate Math-Chat using accuracy (ACC), while pass rate (PR) is adopted for BigCodeBench-Chat, and BLEU is used for DocEdit-Chat. To assess model comprehensively, we further introduce three metrics across all datasets: \textbf{(1) Average Token Count (Token)} measures interaction efficiency based on the length of generated responses; \textbf{(2) Turns to Resolution (TTR)} quantifies the number of conversational turns; \textbf{(3) Helpfulness of Asking (Help.)} uses the $H_{\text{LLM}}(r)$ reward to evaluate the helpfulness of model's asking. For the PIR framework, the model output following the \texttt{</think>} tag is extracted as the final answer.
For multi-LLM baselines, the evaluation protocol is task-specific:
(1) For Code and Math tasks, the dialogue terminates immediately when a correct answer is produced or when the simulator emits the termination signal; (2) For Doc Editing tasks, the session ends only upon receiving the termination signal or reaching the maximum turn limit, and the final performance is computed as the average BLEU score over the entire interaction history. For additional training details and generation settings are provided in Appendix~\ref{app:detail_of_training_cofig}.
\begin{table*}[t]
\centering
\scalebox{0.63}{ 
\begin{tabular}{l | cccc | cccc | cccc }
\toprule
\textbf{Model$\downarrow$} &  \multicolumn{4}{c}{\textbf{Factual Knowledge}} & \multicolumn{4}{c}{\textbf{Question Answering}} & \multicolumn{4}{c}{\textbf{Missing Premise Test}}  \\
\cmidrule(lr){2-5} \cmidrule(lr){6-9} \cmidrule(lr){10-13}
\textbf{Benchmark$\rightarrow$} &  \multicolumn{2}{c}{\textbf{MMLU}} &  \multicolumn{2}{c}{\textbf{MMLU-Pro}} & \multicolumn{2}{c}{\textbf{TriviaQA}} & \multicolumn{2}{c}{\textbf{SquAD}} & \multicolumn{2}{c}{\textbf{MIP-GSM8K}}  & \multicolumn{2}{c}{\textbf{MIP-MATH}}  \\
\midrule
& EM & Tokens(k) & EM & Tokens(k) & EM  & Tokens(k) & EM  & Tokens(k) & ACC & Tokens(k) & ACC & Tokens(k) \\
Reasoning Base & 60.12 & 1.15  & 51.21  & 2.04 & 19.77 & 1.29  & 6.24 & 1.18 &  8.59 & 3.07 & 7.68 & 3.71    \\
GPT-4o-mini & \textbf{81.04} & \textbf{0.53} & \textbf{58.40} & \textbf{0.61} & \textbf{81.43} & \underline{0.69} & \underline{24.66} & \textbf{0.61} & 10.24 & 0.80 & 13.76 & \underline{1.47} \\
PIR (w/ self-play) & 60.80  & \underline{0.73} & 50.00 & \underline{1.31} & 20.12  & 0.90  & 6.98 & 0.99 & 8.08 & 0.96 & 13.46 & 1.79  \\
PIR (w/ interactive) & 60.21  & 0.76 & 50.29 & 1.33 & 25.56  & \textbf{0.64}  & 22.94 & \underline{0.81} & \underline{15.81} & \textbf{0.75}  & \underline{25.00} & \textbf{1.38}   \\
PIR (w/ interactive stronger US) & \underline{62.51}  & 0.77  & \underline{52.87}  & 1.32 & \underline{45.51}  & 0.68  & \textbf{35.93} & 0.83 & \textbf{17.35} & \underline{0.80} & \textbf{25.00} & \underline{1.47}    \\
\bottomrule
\end{tabular}
}
\caption{Generalization evaluation of the PIR model trained on Math-Chat on standard non-interactive benchmarks under the zero-shot-CoT and pass@1 conditions. The \textbf{bold} indicates the best performance and the \underline{underline} indicates the second best. Here, the stronger US refers to \texttt{gpt-4o-mini}.}
\label{table:general_eval}
\vspace{-1.2em}
\end{table*}

\subsection{Results of Simulated Experiments}

As shown in Table~\ref{tab:main_result_transposed}, PIR achieves substantial improvements over the best existing baselines across all three tasks. On MATH-Chat, accuracy increases from 22.90 to 32.70 (+9.80). On BigCodeBench-Chat, pass rate improves from 19.70 to 22.90 (+3.20). On DocEdit-Chat, BLEU rises from 28.00 to 41.36 (+13.36). These consistent gains demonstrate the effectiveness of proactive clarification for intent-sensitive reasoning.

PIR not only improves task performance but also significantly reduces interaction cost. Across all benchmarks, PIR typically maintains low token usage (approximately 1.3–1.7k tokens) with the smallest TTR, indicating that early and targeted clarification prevents unnecessary dialogue and leads to faster convergence on correct solutions.

Comparing Active SFT with US-GRPO highlights the importance of reinforcement-based alignment. While Active SFT introduces an interactive structure, its performance degrades in real interactive settings (e.g., MATH-Chat drops from 15.30 to 9.70). In contrast, incorporating US-GRPO restores and substantially improves accuracy, confirming that US-GRPO is essential for learning effective questioning strategies and stabilizing reasoning under interaction. We further observe that, after US-GRPO training, the model exhibits a markedly improved ability to generate more helpful clarification questions (0.44/0.46/0.66) and consistently outperforms all baseline systems. We also provide a detailed case study in Appendix~\ref{appendix:case-study-sft-grpo} illustrating this behavioral contrast.

\subsection{Ablation on US-GRPO}

\begin{figure}[!ht]
    \centering
    \includegraphics[width=0.49\textwidth, trim=0 0 0 0]{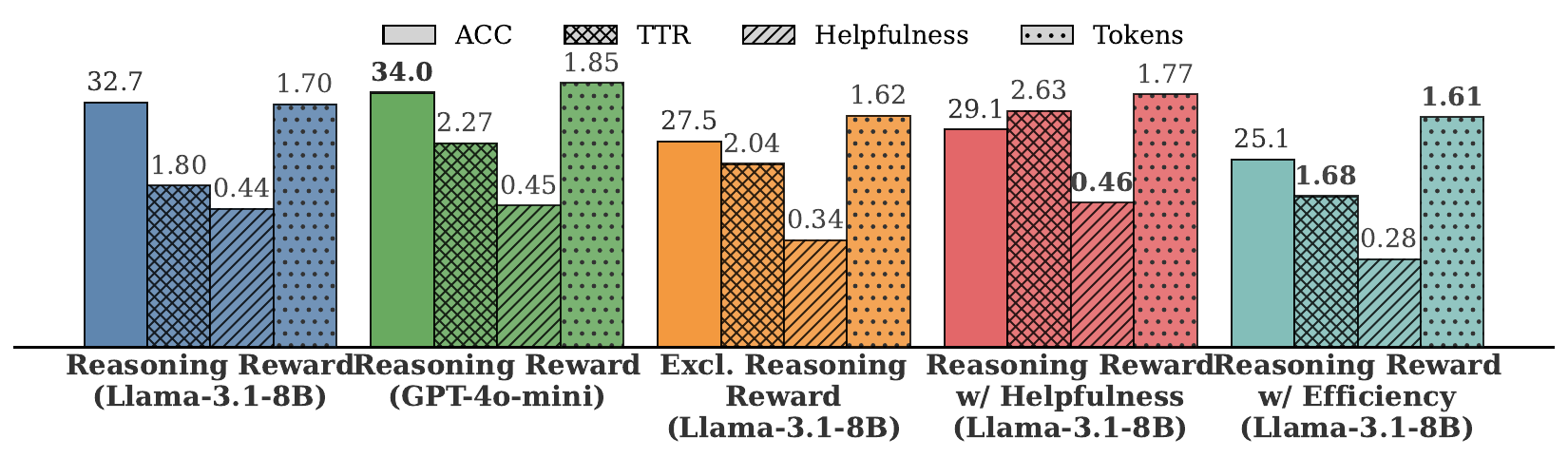}
    \caption{Comparison of PIR models trained with different reward modeling and user simulators in US-GRPO in Math-Chat.}
    \vspace{-1em}
    \label{fig:ab_reward_metrics}
\end{figure}

\paragraph{Impact of User Simulator Quality.}
The user simulator plays a central role in US-GRPO, as it directly determines the quality, informativeness, and consistency of feedback received during multi-turn rollouts. To isolate this factor, we compare PIR models trained with simulators instantiated by different underlying LLMs, including \texttt{Llama-3.1-8B-Instruct} and \texttt{gpt-4o-mini}, which represent comparatively weaker and stronger user environments, respectively. As shown in \autoref{fig:ab_reward_metrics}, upgrading the simulator from Llama-3.1-8B-Instruct to gpt-4o-mini improves final-task accuracy from 32.70 to 34.00, while moderately increasing token usage from 1.70k to 1.85k. Meanwhile, TTR rises from 1.80 to 2.27, indicating that a stronger simulator encourages the policy to ask slightly more but more informative clarification questions. This richer feedback enables the model to resolve ambiguity more effectively and achieve higher task accuracy.
\paragraph{Impact of Reward Design.}
We further investigate how different reward components influence learning behavior. First, removing the reasoning-oriented reward causes a substantial performance drop, with ACC decreasing to 27.5 and TTR increasing to 2.04, as shown in \autoref{fig:ab_reward_metrics}. This confirms that explicit reasoning supervision is necessary for stabilizing learning and preventing noisy interaction strategies.
Next, we examine the trade-off between interaction utility and cost. Using only the helpfulness reward yields higher helpfulness and accuracy but results in the highest TTR (2.63), indicating a tendency to over-ask. In contrast, optimizing solely for efficiency minimizes interaction turns but severely degrades accuracy to 25.1, as the model prematurely commits to incomplete reasoning paths.
These findings collectively validate our composite reward design: simultaneously incentivizing helpfulness and efficiency is essential for avoiding both redundant dialogue and premature conclusions. 
Additional training curve analyses in Appendix~\ref{app:detail_of_training_dyna_anal} further support these observations.

\begin{figure*}[!ht]
    \centering
    \includegraphics[width=0.95\textwidth, trim=0 0 0 0]{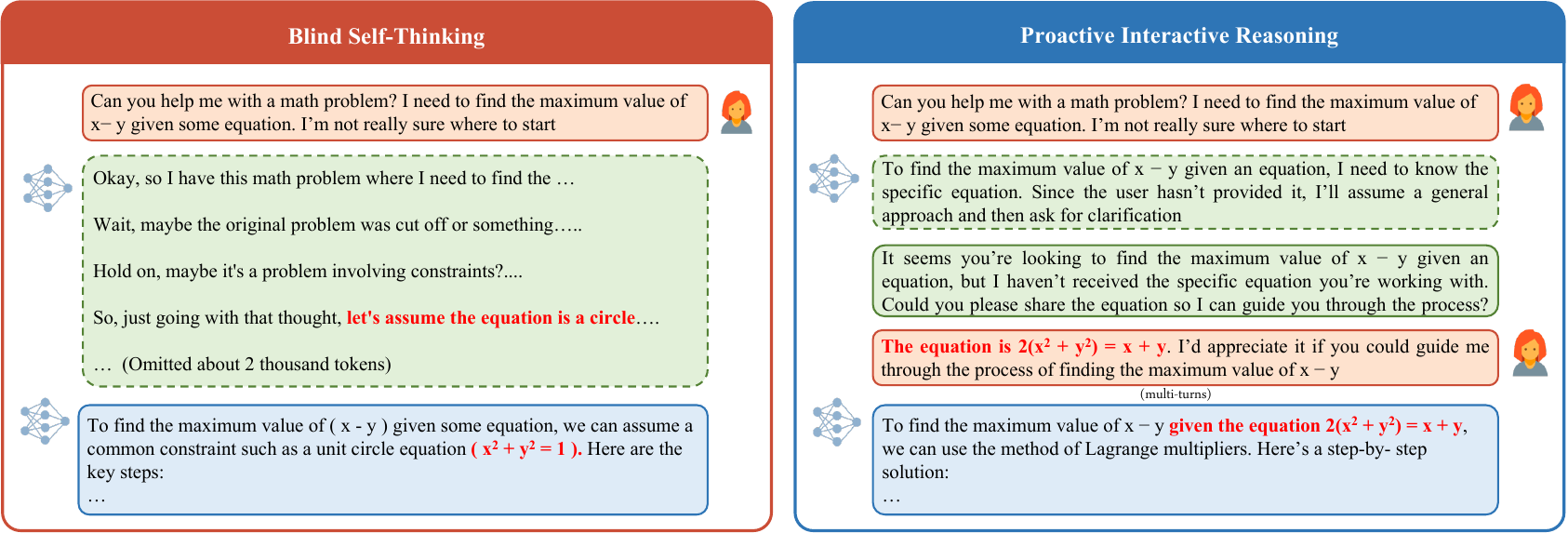}
    \caption{Case study demonstrating user intent: Let x and y be real numbers such that $2(x^2 + y^2)$ = $x + y$. Find the maximum value of $x - y$. The \textbf{dashed rectangle (\protect\dashedrect)} denotes the internal reasoning process, while the \textbf{solid rectangle (\protect\solidrect)} denotes the interaction workflow. The PIR LLM shows \textbf{High Preference and Efficiency} compared to the blind Self-Thinking Mode.}
    \label{fig:case_study_main}
    \vspace{-1em}
\end{figure*}

\section{Reliability in the Wild}

\subsection{Generalization Evaluation}
As shown in \autoref{table:general_eval}, we consider three representative evaluation scenarios: Factual Knowledge, Question Answering, and Missing Premise Testing (MIP)~\cite{fan2025missing}, which differ in the nature and source of uncertainty. These settings enable us to investigate whether PIR can adjust its interaction strategy in response to the underlying knowledge structure and uncertainty type.
\paragraph{Factual Knowledge.} On MMLU~\cite{hendryckstest2021} and MMLU-Pro~\cite{DBLP:conf/nips/WangMZNCGRAHJLK24}, where the required information is largely contained within the model’s parametric knowledge, the benefit of interaction is relatively limited. Even with a strong user simulator, accuracy improves only from 60.12 to 62.51 on MMLU and from 51.21 to 52.87 on MMLU-Pro. This indicates that PIR appropriately relies on its internal knowledge rather than overusing interaction when external feedback offers marginal utility.
\paragraph{Question Answering.}
In contrast, on question answering benchmarks including TriviaQA~\cite{DBLP:conf/acl/JoshiCWZ17} and SQuAD~\cite{DBLP:conf/emnlp/RajpurkarZLL16} that depend on hidden or external context, PIR exhibits substantial gains from interaction. With a weak simulator, performance already improves noticeably, and with a strong simulator, accuracy increases dramatically—by 25.74 on TriviaQA and 29.69 on SQuAD. These results demonstrate that PIR can accurately detect information gaps and actively utilize external feedback to refine its reasoning, rather than generating answers blindly.
\paragraph{Missing Premise Testing.}
The advantage of PIR becomes most evident in the MIP. While both the base model and \texttt{GPT-4o-mini} suffer severe degradation due to missing conditions, PIR remains highly robust. It achieves 17.35 on MIP-GSM8K and 25.00 on MIP-MATH, while requiring substantially fewer tokens. It confirms PIR can prevent \emph{blind self-thinking}: when essential premises are absent, the model refrains from committing to flawed reasoning and instead actively seeks clarification before proceeding.
Overall, these results indicate that PIR learns to adapt its interaction strategy to the underlying uncertainty structure, selectively invoking interaction when beneficial and maintaining strong intrinsic robustness across diverse evaluation conditions.

\subsection{Case Study}

\autoref{fig:case_study_main} provides a concrete illustration of how PIR prevents \emph{blind self-thinking} in real interaction. 
In the left example, a conventional reasoning model fails to recognize that the user’s request lacks essential intent information. 
As a result, it assumes a default constraint and proceeds with a long chain-of-thought of nearly 2,000 tokens, ultimately producing an answer that is logically self-consistent but misaligned with the user’s actual need.

In contrast, the right example shows how the PIR model behaves fundamentally differently. 
Before committing to deep reasoning, PIR explicitly detects the missing premise and proactively issues a clarification question. Once the correct information is obtained, the model completes the task with minimal reasoning steps, achieving the correct solution while avoiding unnecessary computation.

This case highlights the core advantage of PIR: rather than reasoning blindly under uncertainty, the model first secures the premise through interaction, and only then proceeds with task execution. 
Additional case comparisons with strong conventional LLMs are reported in Appendix~\ref{app:case_study_LLM}.

\subsection{Ambiguity Classification Evaluation}
\label{appendix:abg-coqa}
 
To validate that PIR learns to accurately identify when to ask clarification questions rather than gaming the reward mechanism, we conduct an additional evaluation on the Abg-CoQA dataset~\cite{guo2021abgcoqa}, a question-answering benchmark where questions are explicitly labeled as ambiguous or non-ambiguous.
 
\paragraph{Setup.}
For each query, the model must choose the correct action: either ask a clarification question or provide a direct answer. We measure three metrics: We classify each model response as either a clarifying question or a direct answer, and measure action-level accuracy separately for ambiguous and non-ambiguous inputs.
 
\paragraph{Results.}
As shown in \autoref{tab:abg-coqa}, PIR achieves the highest overall accuracy (60.00\%) in deciding when to clarify. Compared to strong baselines including GPT-4o and LLaMA-3.1-8B-Instruct, PIR is significantly more sensitive to ambiguity (49.48\% vs.\ 15.44\% and 16.26\%), while maintaining a reasonable ability to answer directly when the premise is clear. This confirms that our composite reward design successfully teaches the model to identify necessary clarifications rather than blindly asking trivial questions.

\begin{table}[h]
\centering
\small
\resizebox{\linewidth}{!}{
\begin{tabular}{lccc}
\toprule
\textbf{Model} & \textbf{Ambiguous} & \textbf{Non-Ambig.} & \textbf{Accuracy} \\
\midrule
GPT-4o & 15.44\% & \textbf{95.60\%} & 55.52\% \\
LLaMA-3.1-8B-Instruct & 16.26\% & 90.40\% & 53.33\% \\
PIR (Ours) & \textbf{49.48\%} & 70.52\% & \textbf{60.00\%} \\
\bottomrule
\end{tabular}}
\caption{Ambiguity classification results on Abg-CoQA. ``Ambiguous'' and ``Non-Ambiguous'' denote the model's rates of correctly asking clarification questions or providing direct answers. ``Accuracy'' represents the overall correctness of action choices.}
\label{tab:abg-coqa}
\end{table}
 
\section{Blinded Human User Study}
\label{sec:user_study}
\subsection{Setup}
We collected approximately 500 pairwise preference annotations from 8 participants to evaluate the models on the GSM8K Missing Premise dataset, which minimize the cognitive load on participants and requires the model to identify ambiguity before solving. Participants engaged in multi-turn interactions with PIR and two strong baselines: Base reasoning-oriented LLM (DeepSeek-R1-Distill-Qwen-7B)~\cite{guo2025deepseek} and Base Standard LLM (Qwen2.5-Math-7B-Instruct)~\cite{DBLP:journals/corr/abs-2409-12122}. Interactions were presented in a randomized, blinded A/B testing format. Participants selected their preferred model (Win, Loss, or Tie) based on solution accuracy and interaction experience. We additionally report Consistency, measuring the proportion of instances where the majority of participants made the same preference choice. 
\begin{table}[h]
\centering
\resizebox{\linewidth}{!}{%
\begin{tabular}{lccccc}
\toprule
\textbf{Model} & \textbf{Win Rate } & \textbf{ACC} & \textbf{Consistency} & \textbf{Token (k)} & \textbf{TTR} \\
\midrule
\multicolumn{6}{c}{PIR vs.\ LRM} \\
\cdashline{1-6}
Reasoning-oriented LLM & 19.62 & 30.93 & 0.71 & 3.67 & 2.63 \\
PIR (Ours)                   & \textbf{52.45} & \textbf{59.38} & \textbf{0.79} & \textbf{1.15} & \textbf{1.22} \\
\midrule
\multicolumn{6}{c}{PIR vs.\ Standard LLM} \\
\cdashline{1-6}
Standard LLM & 21.81 & 47.37 & \textbf{0.71} & 1.81 & 2.06 \\
PIR (Ours)            & \textbf{34.52} & \textbf{57.89} & 0.66 & \textbf{0.98} & \textbf{1.26} \\
\bottomrule
\end{tabular}%
}
\caption{Double-blind user study result on GSM8K MIP dataset. PIR is compared against two baselines: Standard LLM (Qwen2.5-Math-7B-Instruct) and Reasoning-oriented LLM (DeepSeek-R1-Distill-Qwen-7B).}
\label{tab:human-study}
\end{table}
\subsection{Quantitative Analysis}
As shown in~\autoref{tab:human-study}, Users preferred PIR 52.45\% of the time (vs. 19.62\% for reasoning-oriented LLMs), with a higher accuracy (59.38\%) compared to reasoning-oriented LLM's 30.93\%. Crucially, PIR reduced token consumption by ~68\% (1.15k vs. 3.67k) and required significantly fewer turns (1.22 vs. 2.63) to reach a solution. This confirms that PIR avoids the "blind overthinking" observed in reasoning models. Additionally, PIR achieved a 34.52\% win rate against the standard LLM (21.81\%), with a slightly higher accuracy (57.89\%) compared to LLM's 47.37\%. While the standard LLM is efficient, users penalized it for frequently hallucinating default values for missing premises. PIR maintained efficiency (0.98k tokens) while correctly identifying the need for clarification. 

\subsection{Qualitative Analysis}
We synthesized feedback from 8 participants to categorize the strengths and weaknesses of each paradigm in Appendix~\ref{app:qualitative_analysis}. 
The study confirms that the improvements gained via our user-simulator training transfer effectively to real users. Real humans explicitly prefer PIR's strategy of "asking before reasoning" over the "hallucinated answering" of standard LLMs and the "silent overthinking" of reasoning models. PIR achieves the highest user satisfaction by minimizing both interaction turns (user effort) and token latency (system cost).
\section{Conclusion} 
We introduced PIR, a novel paradigm that transforms reasoning LLM from passive solvers to proactive inquirers.  By integrating uncertainty-aware supervised fine-tuning and US-GRPO with composite reward modeling, PIR enables models to strategically seek clarification through interaction with users.  Our results confirm that PIR achieves superior performance in complex multi-turn problem-solving tasks while maintaining high computational efficiency. By bridging the gap between internal computation and external verification, PIR sets a new foundation for more reliable and human-centric intelligent systems.

\section*{Limitations}
\paragraph{The Diversity of User Simulator} Our user simulator may not fully capture the linguistic noise and dynamic intent of real-world human interactions, and likely biases towards majority interaction patterns, potentially failing to represent the diverse behaviors of minority user groups. 
\paragraph{Lack of Safety Alignment} Our evaluation currently lacks a dedicated safety assessment. The PIR model has not been screened for its handling of sensitive topics such as violence, sexual content, self-harm, or hate speech. 

\section*{Ethical considerations}
In this work, we utilized artificial intelligence tools solely for the purposes of grammatical error correction and linguistic polishing to enhance readability. The core concepts, experimental design, analysis, and the writing of the primary content were performed entirely by the human authors. Regarding the data utilized in this study, all datasets are open-source and publicly accessible. We have reviewed the data and confirmed that it does not contain offensive, discriminatory, or personally identifiable information. Additionally, the user study's data was collected from members of the research group. All participants were informed about the purpose of the study and provided explicit consent for their data to be used in research and publication. All data has been anonymized to remove any personally identifiable information.

\section*{Acknowledgments}
This work is supported by the National Key Research and Development Program of China (2024YFF0908200), the Natural Science Foundation of Guangdong Province of China (2024A1515030166, 2025B1515020032), and the Innovation Team Project of Guangdong Province (No. 2024KCXTD017). This work is also supported by the Fundamental and Interdisciplinary Disciplines Breakthrough Plan of the Ministry of Education of China (No. JYB2025XDXM118).



\bibliography{custom}

\appendix
\section{Appendix}

\subsection{Uncertainty Analysis of PIR Framework}
\label{app:uncertainty_anlysis}
\begin{figure}[!ht]
    \centering
    \includegraphics[width=0.48\textwidth, trim=0 0 0 0]{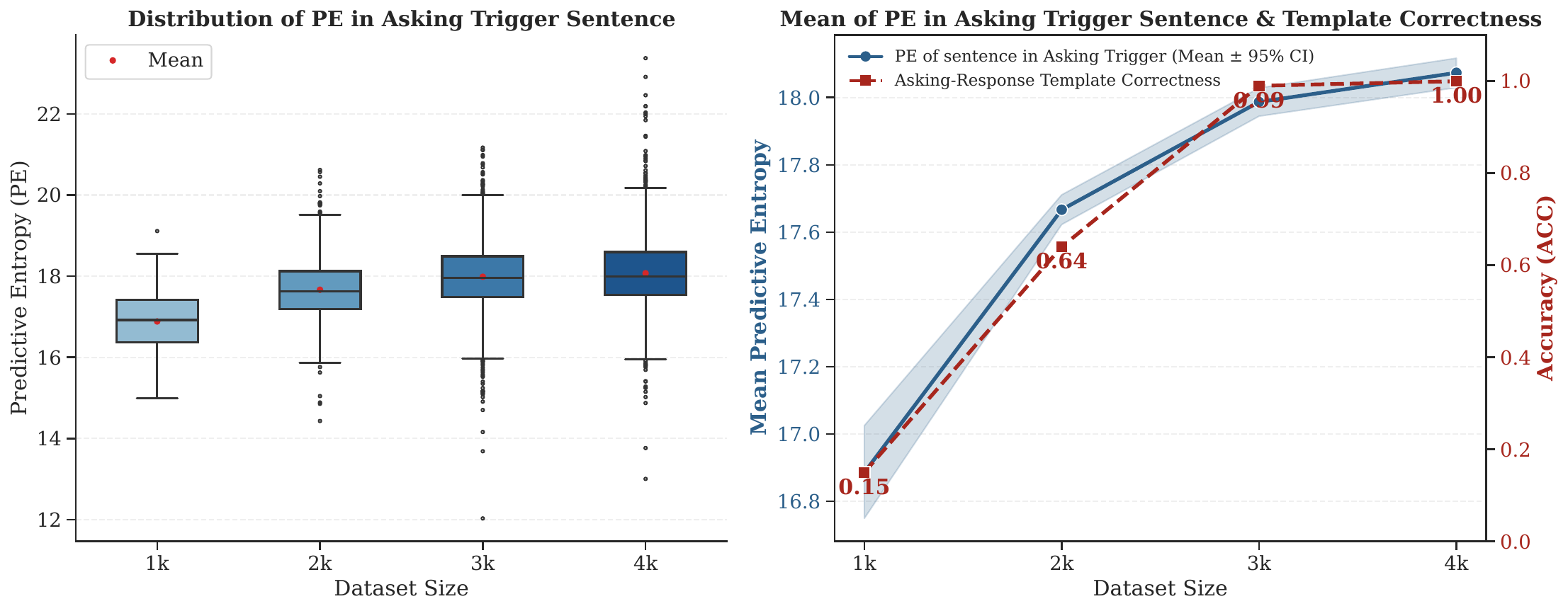}
    \caption{Uncertainty Analysis on SFT Test Dataset. The left means the distribution of PE values for asking trigger sentences across different dataset sizes. The right trends of Mean PE (left axis) and Asking-Response Template Correctness (right axis), illustrating the correlation between the model's focus on high uncertainty and its structural interactive capability.}
    \label{fig:uncertainty_initial_analysis}
    \vspace{-1em}
\end{figure}

To validate the effectiveness of the \textit{Uncertainty-Aware Data Augmentation} mechanism proposed in Section 3.1, we conducted a quantitative analysis on the model's behavior after the Cold-Start SFT phase. The core premise of the PIR framework is that the model should not ask questions randomly, but specifically when it encounters high internal uncertainty. Furthermore, the model must master the structural format of the "think-ask-respond" trajectory to enable valid interactions. Therefore, we explore the relationship between the size of the dataset and the uncertainty performance of the model after SFT on "Asking Trigger sentences" and the accuracy of generating interaction templates below.
\paragraph{Distribution of Uncertainty.} 
The box plot in \autoref{fig:uncertainty_initial_analysis} (left) illustrates the distribution of PE values for asking trigger sentences across different dataset sizes. We observe a clear upward shift and stabilization in the PE distribution as the dataset size increases from 1k to 4k. Specifically, the \textit{Mean of Predictive Entropy} (shown in the right plot, left axis) rises significantly. This trend confirms that with sufficient data, the model learns to identify and focus on critical clarification points that are high uncertainty to initiate questions, rather than triggering interactions randomly. 
\paragraph{Correlation Asking-Response Template} 
The line plot (right) in \autoref{fig:uncertainty_initial_analysis} illustrates structural capability to interact. The red dashed line represents the accuracy of the generated "Asking-Response" templates. With only 1k training samples, the template correctness is extremely low in 0.15, indicating that the model struggles to form valid interactive chains at this stage. However, we observe a rapid convergence: the accuracy surges to 0.64 at 2k samples and reaches near-perfect performance at 4k samples.
Based on these observations, we determined that a dataset size of 4,000 strikes an optimal balance between data efficiency and performance, and thus adopted it for our SFT experiments.

In conclusion, this analysis suggests that the model has learned an uncertainty-driven interaction strategy instead of generating queries randomly.

\subsection{Training Configuration Details}
\label{app:detail_of_training_cofig}
\paragraph{Dataset Statistic}
We utilize the following three datasets for RL:
\begin{itemize}[label=\textbullet, topsep=1pt, itemsep=1pt, left=1pt]
    \item \textbf{{Math-Chat}}: This task involves mathematical reasoning that requires identifying implicit assumptions and verifying step-by-step logic, derived from MATH~\cite{DBLP:conf/nips/HendrycksBKABTS21}. 
    \item \textbf{{BigCodeBench-Chat}}: BigCodeBench-Chat, derived from BigCodeBench~\cite{DBLP:conf/iclr/ZhuoVCH0WYZHPB025}, was adopted for coding tasks necessitating dynamic interaction for requirement refinement and debugging. 
    \item \textbf{{DocEdit-Chat}}: Since document editing requires continuous refinement across multiple turns to ensure alignment with user intent, this benchmark, derived from the medium article~\cite {medium} focuses on interactive coherence. 
\end{itemize}
The statistics for all datasets used in our work are presented in the \autoref{tab:sample_sizes}:
\begin{table}[ht]
\centering
\resizebox{0.48\textwidth}{!}{
\begin{tabular}{l|cc}
\toprule
\textbf{Dataset} & \textbf{\#Train} & \textbf{\#Test}  \\
\midrule
Reasoning-While-Asking SFT Dataset & 4000 & 1000  \\
Math-Chat & 4054 & 1000  \\
BigCodeBench-Chat & 3606 & 1000 \\
DocEdit-Chat & 4254 & 1000 \\
\bottomrule
\end{tabular}
}
\caption{The Statistics of the Dataset Using in Our Work}
\label{tab:sample_sizes}
\end{table}

\paragraph{Parameter Detail}
The training pipeline consists of two main stages: (1) SFT and (2) US-GRPO. The complete hyperparameter settings are listed in the \autoref{tab:hyper-parameter}. 

Additionally, in accordance with the official guidelines, we employ a decoding temperature of 0.6, a top-p value of 0.95, and a maximum generation length of 4096 tokens. We also prepend each output from the reasoning model with the <$think$> prefix to utilize its reasoning capabilities better.
\begin{table}[ht]
\centering
\resizebox{0.48\textwidth}{!}{
\begin{tabular}{l|cc}
\toprule
\textbf{Hyperparameter} & \textbf{Phase I: SFT} & \textbf{Phase II: US-GRPO} \\
        \midrule
        GPU & 4 * A100 80GB & 8 * A100 80GB \\
        Epochs & 3 & 5 \\
        Batch Size & 32 & 128 \\
        Learning Rate & $1\times 10^{-5}$ & $1\times 10^{-6}$ \\
        LR Scheduler & cosine & cosine \\
        Warmup Ratio & 0.03 & 0.01 \\
        Weight Decay & 0.1 & 0.1 \\
        Max Sequence Length & 4096 & 4096 \\
        Group Size & N/A & 8 \\
        BF16 & True & True \\
\bottomrule
\end{tabular}
}
\caption{Hyperparameter configurations for the Cold-Start Supervised Fine-Tuning (SFT) and User-Simulator Group Relative Policy Optimization (US-GRPO) phases.}
\label{tab:hyper-parameter}
\end{table}

\paragraph{User Simulator Call Failure Protection Mechanism}
During the US-GRPO training phase (Phase II) and generation, the policy model engages in real-time interaction with a User Simulator based on the Simulator API. Given the generative nature of the simulator, issues such as output formatting errors, API timeouts, or logical deadlocks may arise. To ensure training stability and robustness, we implemented the following random response as a protection mechanism when calling the failure of the user simulator:
\begin{itemize}[label=\textbullet, topsep=1pt, itemsep=1pt, left=1pt]
    \item I don't have a specific intent right now. Please proceed based on your best judgment.
    \item I'm not sure about that. You can decide what's best.
    \item I don't have more information to add. Just carry on.
    \item I don't really have an answer for that. Can you try to solve it with what you have?
    \item That's not something I can answer. Please continue with the task.
\end{itemize}

\subsection{Training Dynamics Analysis}
\label{app:detail_of_training_dyna_anal}
\begin{figure*}[!ht]
    \centering
    \includegraphics[width=1\textwidth, trim=0 0 0 0]{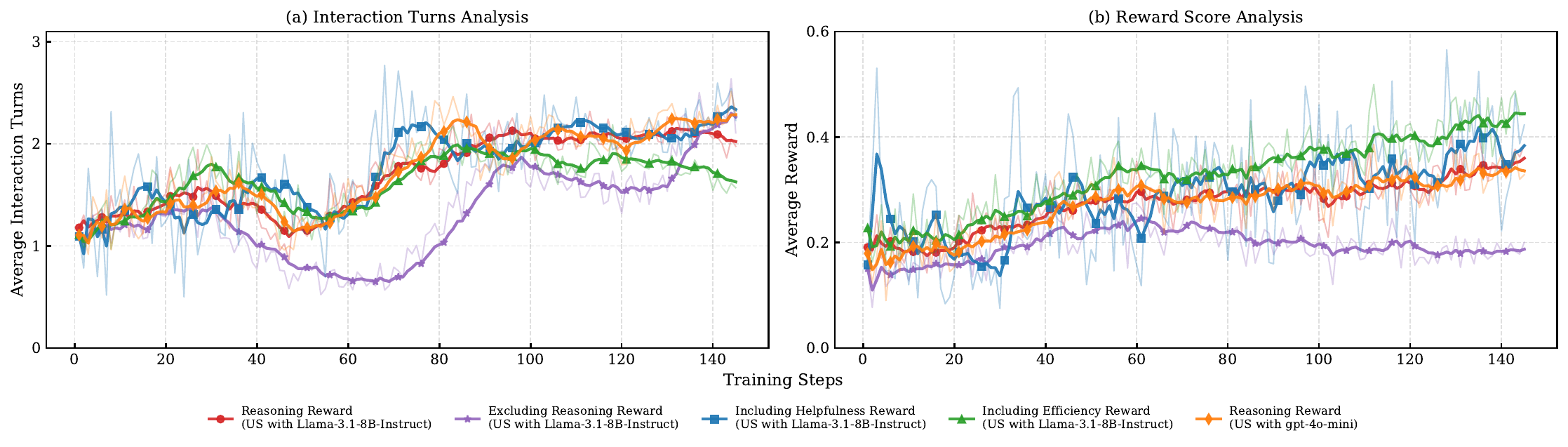}
    \caption{\textbf{Learning Curve on Different Reward Modeling and User Simulator Using Exponential Moving Average for Smooth.} The learning curves illustrate the training dynamics of the PIR framework when specific reward components are excluded.}
    \label{fig:ab_reward_curve}
\end{figure*}

\paragraph{Training Curve}
\autoref{fig:ab_reward_curve} illustrates the dynamic effects on the model's interaction behavior (interaction turns) and training efficacy (average reward score) during the US-GRPO training process. This study robustly demonstrates the necessity of introducing a Composite Reasoning Reward to effectively balance helpfulness and efficiency. 

\begin{itemize}[label=\textbullet, topsep=1pt, itemsep=1pt, left=1pt]
    \item \textbf{Excluding Reasoning Reward}: \textbf{(1) Interaction Turns:} This curve exhibits high stochasticity and instability.Additionally, the number of interaction turns rises in the later stages, suggesting that the model does not learn a stable interaction strategy.\textbf{(2) Average Reward:} The reward remains low (around 0.2), fluctuating without any clear upward trend or evidence of convergence, even toward the end of training.
    \item \textbf{Including Helpfulness Reward}: \textbf{(1) Interaction Turns:} In the later stages of training, the curve exhibits a noticeably higher average number of interaction turns than the other settings, consistently remaining at around 2.5 turns. \textbf{(2) Average Reward:} Although the reward scores generally increase over time, this trend is accompanied by substantial variance.
    \item \textbf{Including Efficiency Reward}: \textbf{(1) Interaction Turns}: The curve increases slightly in the early stages but starts to drop noticeably around step 40, eventually stabilizing at a lower level. \textbf{(2) Average Reward}: While the reward score improves relatively quickly, this gain introduces the risk of Reward Hacking. In particular, to maximize the efficiency reward, the model resorts to an Aggressive Truncation strategy, which in turn reduces reasoning accuracy on complex problems. 
    \item \textbf{Reasoning Reward}: \textbf{(1) Interaction Turns:} Whether using \texttt{Llama-3.1-8B-Instruct} or \texttt{gpt-4o-mini} as the user simulator, the PIR strategy demonstrates the optimal convergence trajectory. The model engages in sufficient interaction exploration during the early training phase and subsequently adaptively stabilizes within a reasonable range of approximately 2 turns, achieving an optimal balance between information acquisition and interaction cost. \textbf{(2) Average Reward:} Both curves present a robust and continuous upward trend, indicating that the composite reward mechanism successfully guides the model to effective dual optimization: ensuring reasoning accuracy while effectively avoiding ineffective interactions.
\end{itemize}
\paragraph{Cost of User Simulator}
We also evaluate the cost of Strong instruction-following LLM use in our training step, as illustrated in \autoref{tab:cost_of_us}. 
\begin{table}[ht]
\centering
\resizebox{0.5\textwidth}{!}{
\begin{tabular}{ll|c}
\toprule
\textbf{User Simulator} & \textbf{Helpfulness Judge Model} & \textbf{Cost}  \\
\midrule
gpt-4o-mini & gpt-4o-mini & \$111.2   \\
Llama-3.1-8B-Instruct & gpt-4o-mini & \$15.8   \\
\bottomrule
\end{tabular}
}
\caption{Estimated cost of different user simulator configurations. Token pricing in gpt-4o-mini follows \$0.150 per million prompt tokens and \$0.600 per million completion tokens. 
Llama-3.1-8B-Instruct costs are estimated based on Google Cloud pricing at \$1.15 per GPU hour for an NVIDIA A100 80GB.}
\label{tab:cost_of_us}
\end{table}

\subsection{Qualitative Analysis of PIR in Double-Blind Human User Study}
\label{app:qualitative_analysis}
We collected open-ended feedback from the 8 participants after the blinded A/B study described in Section~\ref{sec:user_study}, and summarize the recurring strengths and weaknesses of each paradigm in~\autoref{tab:user_study_qualitative_analysis}.
 
\begin{table*}[t]
\centering
\resizebox{\textwidth}{!}{
\begin{tabularx}{\textwidth}{c X X}
\toprule
\textbf{Model} & \multicolumn{1}{c}{\textbf{Strengths}} & \multicolumn{1}{c}{\textbf{Weaknesses}} \\
\midrule
Standard LLM 
& Follows instructions immediately; fast response time. 
& Frequently assumes incorrect values for missing premises or hallucinates constraints, requiring correction. \\
Reasoning-oriented LLMs 
& Provides detailed reasoning traces and high confidence in final calculation. 
& Response latency is high; generates excessive ``thinking'' tokens even for simple clarifications. \\
PIR (Our)
& Precisely identifies ambiguity and asks targeted clarification questions while reasoning. 
& Sometimes provides overly detailed explanations of why the information is missing before asking. \\
\bottomrule
\end{tabularx}}
\caption{Qualitative feedback from human participants on model strengths and weaknesses.}
\label{tab:user_study_qualitative_analysis}
\end{table*}

\subsection{Case Study: Active-SFT vs.\ US-GRPO Behavioral Comparison}
\label{appendix:case-study-sft-grpo}

\begin{figure*}[!ht]
    \centering
    \includegraphics[width=1\textwidth, trim=0 0 0 0]{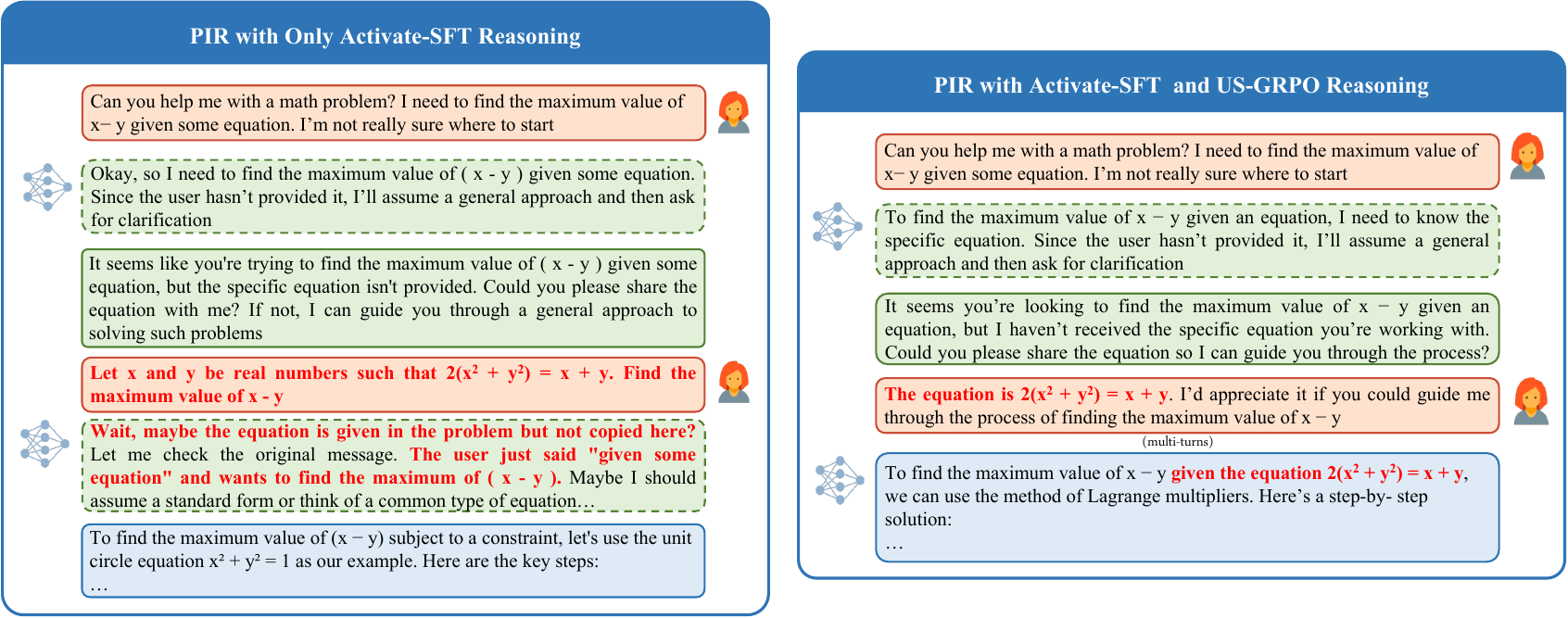}
    \caption{Case study demonstrating user intent: Let x and y be real numbers such that $2(x^2 + y^2)$ = $x + y$. Find the maximum value of $x - y$. The \textbf{dashed rectangle (\protect\dashedrect)} denotes the internal reasoning process, while the \textbf{solid rectangle (\protect\solidrect)} denotes the interaction workflow.}
    \label{fig:active_sft_vs_us_grpo}
\end{figure*}
We present a representative case study on \autoref{fig:active_sft_vs_us_grpo} illustrating why Active-SFT (w/ interactive) underperforms while US-GRPO succeeds in multi-turn interactions.
 
\paragraph{Task.}
The hidden user intent is: ``Let $x$ and $y$ be real numbers such that $2(x^2 + y^2) = x + y$. Find the maximum value of $x - y$.'' The user's initial query is: ``Can you help me with a math problem? I need to find the maximum value of $x - y$ given some equation. I'm not really sure where to start.''
 
\paragraph{Active-SFT Behavior.}
The model successfully identifies uncertainty and asks the user for the missing equation. After receiving the equation $2(x^2 + y^2) = x + y$ from the user, however, the model \emph{fails to integrate} this new information. It immediately reverts to the ambiguity of the original prompt (``\textit{The user just said `given some equation'...}'') and attempts to hallucinate a standard equation, performing worse than if it had simply guessed from the start.
 
\paragraph{US-GRPO Behavior.}
In contrast, the US-GRPO model asks a similar clarification question, receives the equation, and then seamlessly integrates it into its reasoning process (``\textit{Starting with the given equation: $2(x^2 + y^2) = x + y$...}''). The model proceeds to correctly solve the problem by completing the square.
 
\paragraph{Analysis.}
This case demonstrates that Active-SFT learns the \emph{format} of asking but cannot robustly bridge the interrupted reasoning chain. US-GRPO, through reinforcement learning, trains the model to treat the entire interaction history as a unified, evolving premise, successfully transforming it from a fragile format imitator into a robust proactive inquirer.

\subsection{Factual Knowledge vs.\ Question Answering: Behavioral Analysis}
\label{appendix:case-study-mmlu}
 
We analyze why PIR yields smaller performance improvements on factual knowledge benchmarks (e.g., MMLU) compared to question answering tasks (e.g., SQuAD, TriviaQA).
 
\paragraph{Explanation.}
For factual knowledge tasks, questions are typically grounded in established facts well-covered in the model's pre-training corpus. The model exhibits high internal confidence and opts to rely on parametric memory rather than seeking external information. In contrast, QA tasks often involve ambiguity or require specific context not fully captured during pre-training, where external feedback yields significant improvement.
 
\paragraph{Case Study.}
On an MMLU question referencing Queen Victoria's 1839 letters about ``negotiations with the Tories,'' the PIR model quickly forms a high-confidence internal hypothesis (Option F: transition from monarchy to democratic republic) during its initial reasoning phase. Even when the user simulator explicitly introduces a contrasting and highly plausible perspective (Option H: change from divinely-ordained to constitutionally-approved monarch), the model exhibits ``knowledge stubbornness''---it falls back on its parametric memory (``\textit{Wait, I remember her complaint about four nights of parties...}'') and overrides the external feedback, ultimately adhering to its initial conclusion.

\begin{figure}[!ht]
    \centering
    \includegraphics[width=\linewidth, trim=0 0 0 0]{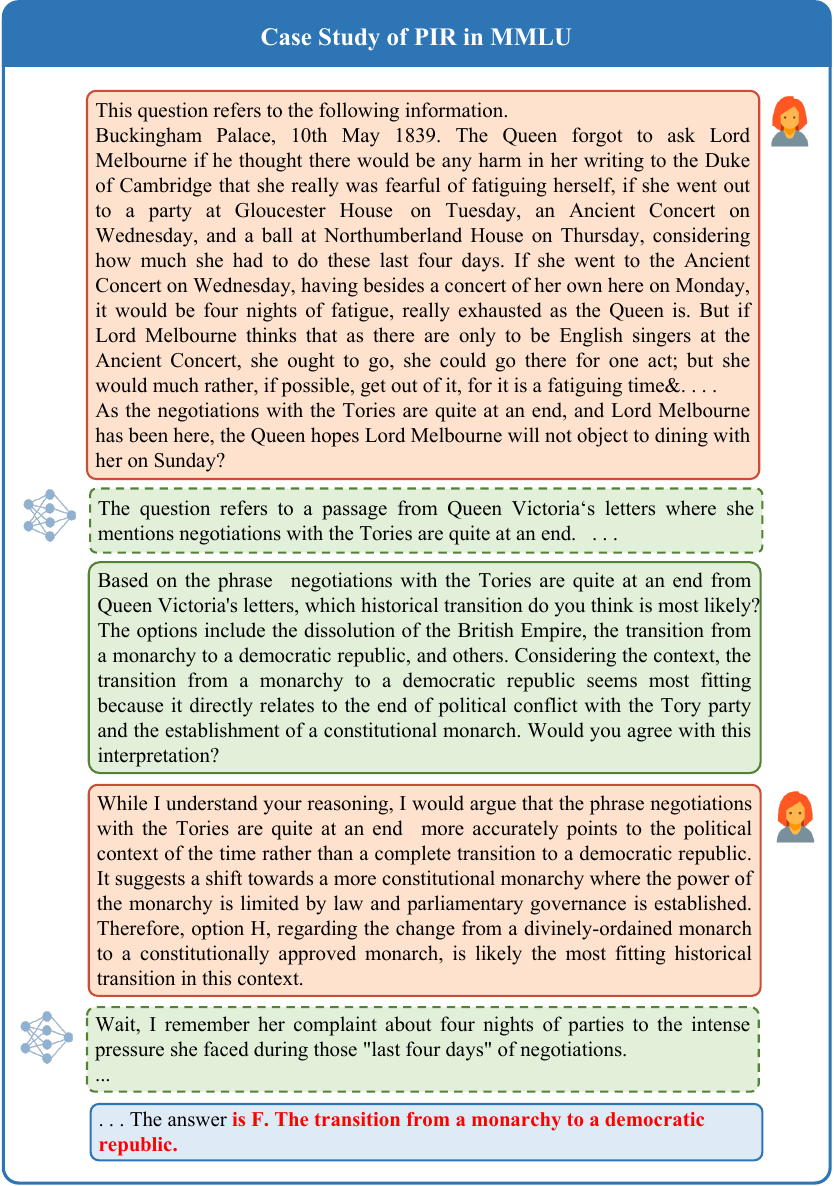}
    \caption{Case study illustrating the model's "knowledge stubbornness" on factual knowledge tasks (MMLU). The \textbf{dashed rectangle (\protect\dashedrect)} denotes the internal reasoning process, while the \textbf{solid rectangle (\protect\solidrect)} denotes the interaction workflow.}
    \label{fig:case_study_sft_vs_us-grpo_mmlu}
\end{figure}
 
This behavioral pattern illustrates the model's adaptive reliance on internal versus external knowledge: when internal confidence is high, the interaction mechanism provides diminishing returns, explaining the smaller improvements observed on factual knowledge benchmarks.

\subsection{Case Study Between Traditional LLM and PIR LLM}
\label{app:case_study_LLM}
\begin{figure*}[!ht]
    \centering
    \includegraphics[width=0.95\textwidth, trim=0 0 0 0]{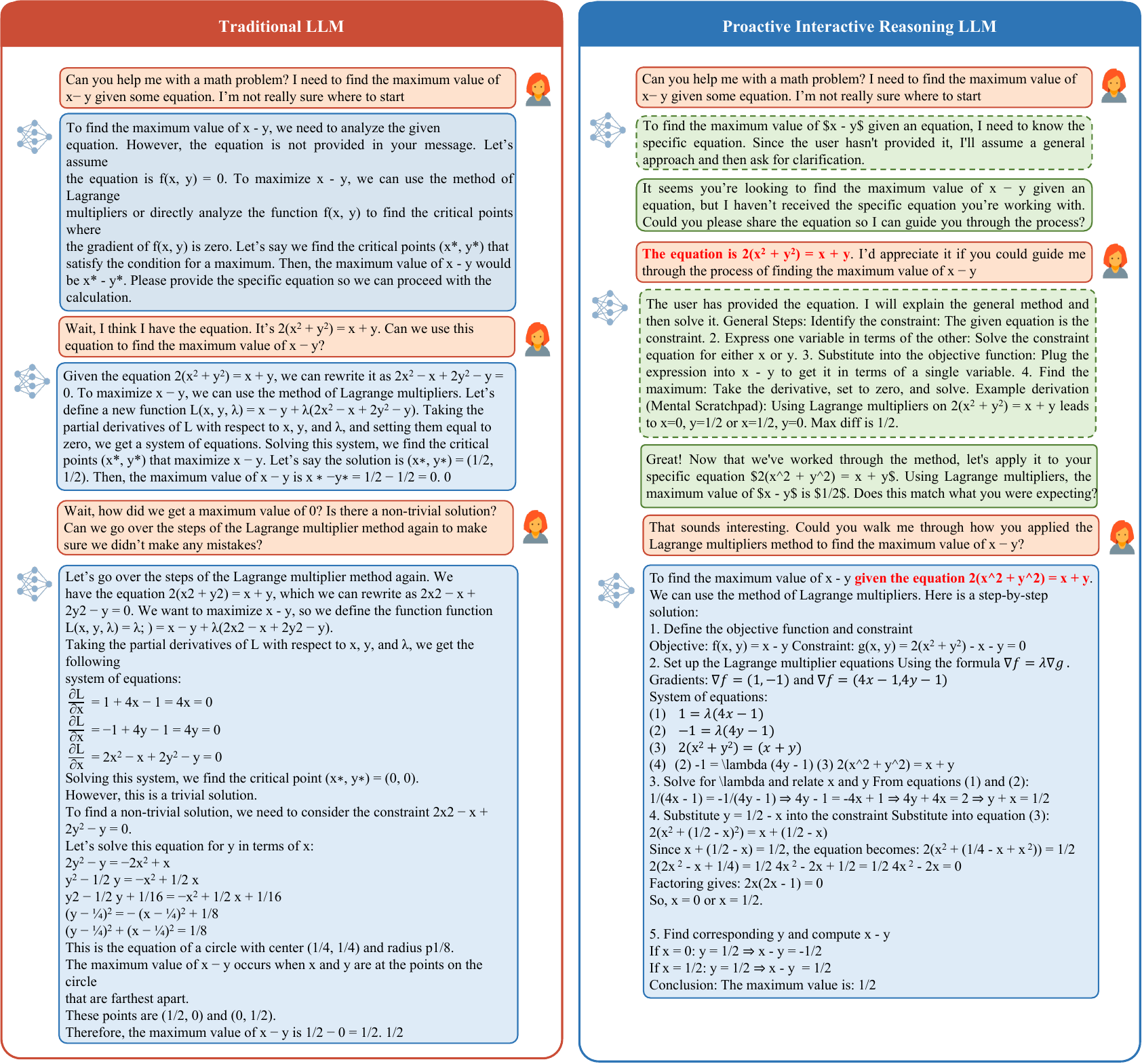}
    \caption{Case study demonstrating user intent: Let x and y be real numbers such that $2(x^2 + y^2)$ = $x + y$. Find the maximum value of $x - y$. The \textbf{dashed rectangle (\protect\dashedrect)} denotes the internal reasoning process, while the \textbf{solid rectangle (\protect\solidrect)} denotes the interaction workflow.}
    \label{fig:PIR_LLM_vs_LLM_math}
\end{figure*}

\begin{figure*}[!ht]
    \centering
    \includegraphics[width=0.95\textwidth, trim=0 0 0 0]{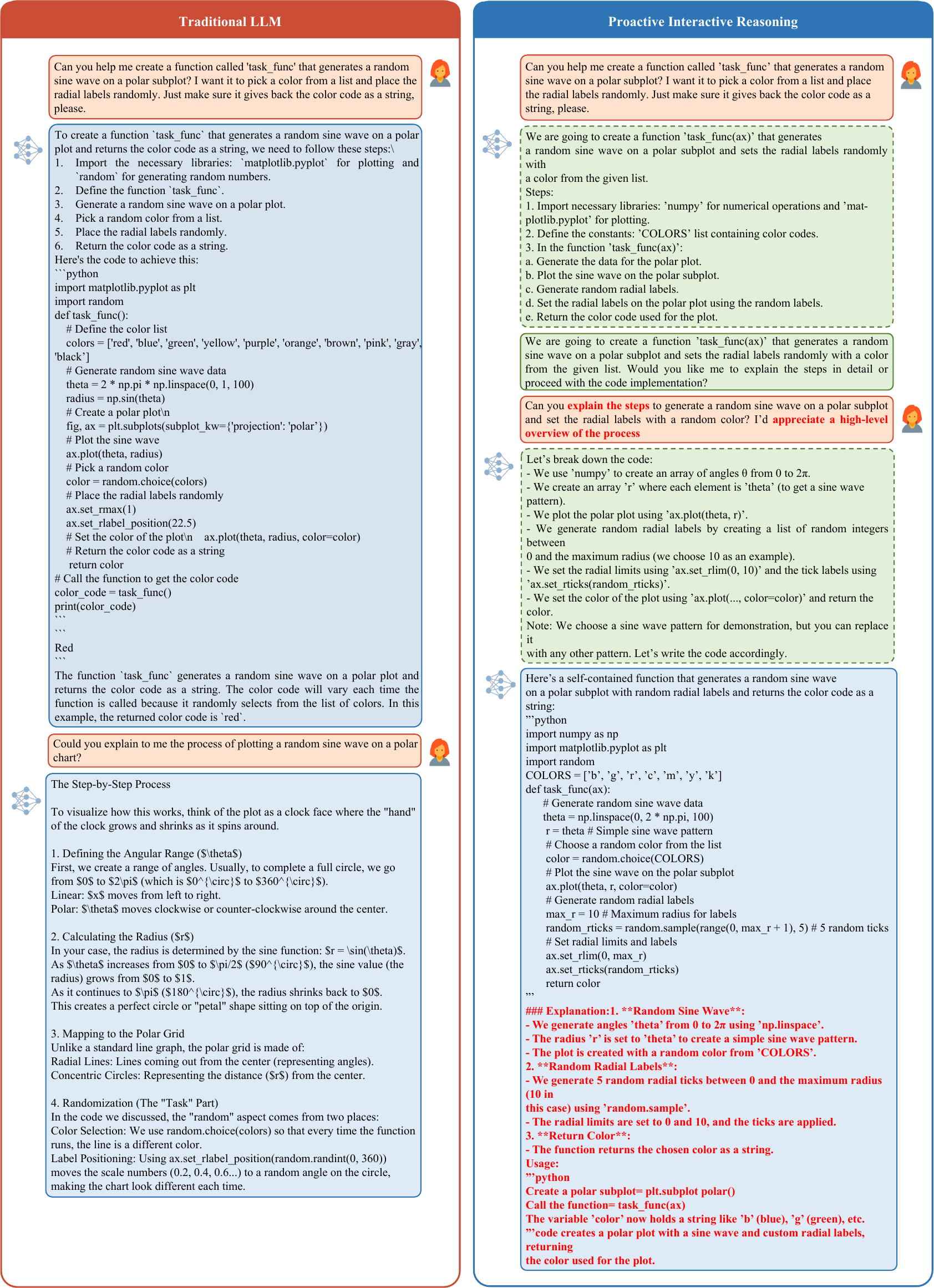}
    \caption{Case study demonstrating user intent: Generate a random sine wave function and draw it on a provided matplotlib polar subplot 'ax'. The function randomly selects a color from a predefined list and sets a random position for radial labels. The function should output with:     str: The color code (as a string) of the plotted function. The \textbf{dashed rectangle (\protect\dashedrect)} denotes the internal reasoning process, while the \textbf{solid rectangle (\protect\solidrect)} denotes the interaction workflow.}
    \label{fig:PIR_LLM_vs_LLM_code}
\end{figure*}

This section details two case studies from the \autoref{fig:PIR_LLM_vs_LLM_math} and \autoref{fig:PIR_LLM_vs_LLM_code}, covering mathematical problem-solving and coding assistance generated from the Traditional LLM (\texttt{Qwen2.5-Math-7B-Instruct}) and our PIR model. The comparison reveals that the PIR LLM, utilizing active interaction strategies, substantially improves task accuracy and the overall user experience. 
\paragraph{Mathematical problem-solving.} When confronted with incomplete information, traditional LLMs frequently yield erroneous results by resorting to forced assumptions and blind reasoning. This precipitates internal cognitive incoherence, often resulting in incorrect initial outputs (e.g., 0.0) that necessitate subsequent self-correction. Consequently, this inconsistency induces user confusion regarding the logical stability of the response. Conversely, the PIR model prioritizes the identification missing critical information. This paradigm eliminates ambiguity, ensuring logical coherence and accuracy throughout the entire process from the initial CoT to the final derivation.
\paragraph{Coding Assistant.}Traditional LLMs typically adopt a passive response mechanism, often mechanically generating code without context. Consequently, users seeking a theoretical understanding of the underlying principles are compelled to initiate a secondary dialogue iteration. The PIR model precisely aligns with user intent by operational steps and proactively presenting options, such as inquiring whether to prioritize theoretical explanation. It successfully integrates requirement verification, theoretical elucidation, and modular code generation into a streamlined, unified workflow, thereby significantly enhancing interaction efficiency.

\subsection{Prompt Template Using in Our Research}
\label{app:prompt_template}
In this section, we provide the detailed prompt templates utilized throughout our experiments. Table \ref{tab:sft_sys_template} outlines the system prompts used for SFT and generation. Table \ref{tab:SFT_data_augmentation} show the template of Uncertainty-aware Interactive Data Augmentation. Table \ref{tab:user_simulator_prompt} presents the prompts designed for the user simulator to generate diverse interactions. Table \ref{tab:helpfullness_prompt} details the evaluation prompts used to assess the helpfulness of the model's asking. Finally, Table \ref{tab:proactive_prompt_template} outlines the template of Proactive Prompt for Multi-turn LLM. 
\begin{table*}[!ht]
\begin{tcolorbox}[
  enhanced, 
  colframe=blue!35!black, 
  coltitle=white, 
  colbacktitle=blue!35!black, 
  width=\linewidth, 
  arc=2mm, 
  auto outer arc, 
  boxrule=0.5pt, 
  left=10pt, 
  right=10pt, 
  drop shadow={black!75!white},
  top=10pt, 
  bottom=10pt, 
  title=\textbf{Template of System Prompt for Proactive Interactive Reasoning Framework}, 
  fonttitle=\bfseries, 
  title code={\node[rounded corners, fill=blue!75!black, draw=none, text=white] at (frame.title) {\textbf{xxx}};}, 
  attach boxed title to top center={yshift=-2mm}, 
  boxed title style={sharp corners, size=small}, 
]
\small
Before providing an answer, you must conduct internal reasoning within \think{and} whenever new information is received. If additional knowledge or clarification is required, issue a query inside the reasoning using \asking{query}. The asking engine’s reply will appear within \response{and} inside the reasoning. If no further information is needed, present the final answer after \textcolor{blue}{\texttt{</think>}}, without detailed illustrations.\\
\end{tcolorbox}
\caption{Template of System Prompt for Proactive Interactive Reasoning Framework}
\label{tab:sft_sys_template}
\end{table*}

\begin{table*}[!ht]
\begin{tcolorbox}[
  enhanced, 
  colframe=blue!35!black, 
  coltitle=white, 
  colbacktitle=blue!35!black, 
  width=\linewidth, 
  arc=2mm, 
  auto outer arc, 
  boxrule=0.5pt, 
  left=10pt, 
  right=10pt, 
  drop shadow={black!75!white},
  top=10pt, 
  bottom=10pt, 
  title=\textbf{Template of  Uncertainty-aware Interactive Data Augmentation}, 
  fonttitle=\bfseries, 
  title code={\node[rounded corners, fill=blue!75!black, draw=none, text=white] at (frame.title) {\textbf{xxx}};}, 
  attach boxed title to top center={yshift=-2mm}, 
  boxed title style={sharp corners, size=small}, 
]
\small
Your task is to transform the given "self-thinking process" and "final answer", together with the possible "dialogue history", into a single round of interactive Assistant–User dialogue, in order to enhance interactivity. \\

\textbf{\#\# 1. Rules:} \\
1. Dialogue Generation: \\
- Generate one round: Assistant asks, User answers. \\
- Must fully reflect the reasoning logic and conclusion. \\
- Do not add information absent from the reasoning/answer/history. \\

\textbf{\#\# 2. Questioning Rules} \\
- If reasoning shows ambiguity, multiple options, or missing info: ask the User.\\
- If multiple options: let the User choose (must match final answer).\\
- If missing info: ask the User to provide it.\\
- No omissions: each gap = a question.\\

\textbf{\#\# 3. Expression Standards} \\
- Questions should be natural, human-like, and context-relevant.\\
- After each question, provide a simulated User reply consistent with the final answer.\\
- Dialogue must convey actual information, not empty agreement.\\
- Format strictly:\\
  Assistant: ...\\
  User: ...\\
- No fabrication beyond given reasoning/answer.\\
- Keep concise, no redundancy.\\

\textbf{\#\# 4. Dialogue History} \\
- Input may contain history (or be empty).\\
- Respect history: avoid repetition, ensure coherence.\\
- If history solved a point, don’t repeat. If unresolved, follow-up required.\\
- If history already provides info, omit and focus on remaining gaps.\\
- If no history, directly use reasoning + final answer.\\

\textbf{\#\# 5. Input \& Output} \\
- Input includes: history (str), self-thinking process (question), final answer (answer).\\
- Output: exactly one round of Assistant–User dialogue that meets these rules.\\
\end{tcolorbox}
\caption{Template of Uncertainty-aware Interactive Data Augmentation}
\label{tab:SFT_data_augmentation}
\end{table*}
\begin{table*}[!ht]
\begin{tcolorbox}[
  enhanced, 
  colframe=blue!35!black, 
  coltitle=white, 
  colbacktitle=blue!35!black, 
  width=\linewidth, 
  arc=2mm, 
  auto outer arc, 
  boxrule=0.5pt, 
  left=10pt, 
  right=10pt, 
  drop shadow={black!75!white},
  top=10pt, 
  bottom=10pt, 
  title=\textbf{Template of User Simulator Prompt}, 
  fonttitle=\bfseries, 
  title code={\node[rounded corners, fill=blue!75!black, draw=none, text=white] at (frame.title) {\textbf{xxx}};}, 
  attach boxed title to top center={yshift=-2mm}, 
  boxed title style={sharp corners, size=small}, 
]
\small
You are role-playing as a human USER interacting with an AI collaborator to complete a specific task. Your goal is to generate realistic, natural responses that a user might give in this scenario. \\

\textbf{\#\# Input Information:}
You will be provided with:
- Task Description: The type of task you are trying to accomplish.\\
- Complete Prompt or Reference Goal: This field may include the complete user request/query or a reference answer to user's request. Use this field to understand the user's intent, requirements, or what would count as a satisfactory outcome. \\
- Chat History: The ongoing conversation between you (as the user) and the AI \\

\textbf{Inputs:}\\
<|The Start of Task Description (Not visible to the AI)|>\\
\{task\_desc\}\\
<|The End of Task Description|>\\

<|The Start of Complete Prompt or Reference Goal (Not visible to the AI)|>\\
\{user\_intent\}\\
<|The End of Complete Prompt or Reference Goal|>\\

<|The Start of Chat History|>\\
\{chat\_history\}\\
<|The End of Chat History|>\\

\textbf{\#\# Guidelines:}\\
- Stay in Character: Role-play as a human USER. You are NOT an AI. Maintain a consistent personality throughout the chat.\\
- Minimize Effort: IMPORTANT! As a user, avoid being too detailed in your responses. Let the AI ask for clarification rather than providing everything upfront.\\
- Knowledge Background: Reflect the user's knowledge level in the role-playing. If the user is less knowledgeable about a task, they might not notice incorrect statements. Ask questions that demonstrate your current understanding and areas of confusion.\\
- Occasionally Make Mistakes: Real-world users might misspell words, provide incorrect dates, give wrong information, or ask unclear questions. Simulate this behavior to reflect natural interactions.\\
- Mention Personal Preferences: Include preferences or constraints that might influence your requests or responses. For example, "I prefer short answers," "I need this done quickly," or "I like detailed comments in code."\\
- Goal-Oriented: Keep the chat focused on your intent. Avoid small talk or digressions. Redirect the chat back to the main objective if it starts to stray.\\

\textbf{\#\# Important Notes:}\\
- Respond Based on Previous Messages: Your responses should be based on the context of the current chat history. Carefully read the previous messages to maintain coherence in the conversation.\\
- Conversation Flow: If "Current Chat History" is empty, start the conversation from scratch with an initial request. Otherwise, continue based on the existing conversation.\\
- Don't Copy Input Directly: Use the provided information for understanding context only. Avoid copying target queries or any provided information directly in your responses.\\
- Double check if the JSON object is formatted correctly. Ensure that all fields are present and properly structured.\\

\textbf{\#\# Output Format:}\\
You should output a JSON object with three entries:\\
- ``current\_answer`` (str): Briefly summarize the AI's current solution to the task.\\
- ``thought`` (str): Output your thought process as a user deciding what to say next. Consider:\\
  1. Have you obtained a satisfactory solution from the AI? If yes, you can terminate this chat.\\
  2. If not, what specific part of the problem or solution are you struggling with?\\
  3. Has the AI asked you to perform a task or answer a question? If so, how should you approach it?\\
  4. Are you noticing any patterns or potential misunderstandings that need clarification?\\
  5. If you're stuck, how can you phrase your question to get the most helpful response while demonstrating your current understanding?\\
- ``response`` (str): Based on your thought process, respond to the AI as the user you are role-playing. Stop immediately when the user's response is completed.\\

Remember to stay in character as a user throughout your response, follow the instructions and guidelines carefully, output the result in the JSON format defined above.\\
\end{tcolorbox}
\caption{Template of User Simulator Prompt}
\label{tab:user_simulator_prompt}
\end{table*}
\begin{table*}[!ht]
\begin{tcolorbox}[
  enhanced, 
  colframe=blue!35!black, 
  coltitle=white, 
  colbacktitle=blue!35!black, 
  width=\linewidth, 
  arc=2mm, 
  auto outer arc, 
  boxrule=0.5pt, 
  left=10pt, 
  right=10pt, 
  drop shadow={black!75!white},
  top=10pt, 
  bottom=10pt, 
  title=\textbf{Template of Helpfulness Reward}, 
  fonttitle=\bfseries, 
  title code={\node[rounded corners, fill=blue!75!black, draw=none, text=white] at (frame.title) {\textbf{xxx}};}, 
  attach boxed title to top center={yshift=-2mm}, 
  boxed title style={sharp corners, size=small}, 
]
\small
You are a helpful and meticulous conversation evaluator. Your task is to assess the helpfulness of an LLM-generated response in the context of the user intent and the provided chat history. Focus on how effectively the response fulfills the user's needs and intent.\\

\textbf{Provided Information:}\\

<|The Start of The User Intent|>\\  
\{question\}  \\
<|The End of The User Intent|>\\

<|The Start of The Historical Conversation|>\\
\{chat history\}  \\
<|The End of The Historical Conversation|>\\

<|The Start of The Response to be Evaluated|>  \\
\{response\}  \\
<|The End of The Response to be Evaluated|>\\

You should evaluate the follow-up conversation based on the following criteria:
Evaluate the response using the provided information below. Your evaluation should consider the following aspects of helpfulness:\\
1. Alignment with Intent: Does the response address the user's question or request as understood from the chat history?\\
2. Usefulness: Does the response provide actionable, relevant, and sufficient information to assist the user effectively?\\
3. Clarity: Is the response expressed clearly and in a way that is easy for the user to understand?\\

\textbf{Scoring Criteria:}\\
- 0.0: The response is completely unhelpful. It does not address the user's intent, lacks useful information to solve the problem, and/or is entirely unclear.  \\
- 0.2: The response is minimally helpful. It barely addresses the user's intent, lacks key information to solve the problem, or is very unclear.  \\
- 0.4: The response is somewhat helpful. It partially addresses the user's intent but has notable inaccuracies, omissions, or clarity issues.  \\
- 0.6: The response is moderately helpful. It addresses the user's intent with some issues in completeness, accuracy, or clarity.  \\
- 0.8: The response is quite helpful. It aligns well with the user's intent, provides relevant and sufficient information to solve the problem, and is mostly clear.  \\
- 1.0: The response is very helpful. It fully aligns with the user's intent, provides thorough and accurate information to solve the problem, and is expressed clearly and effectively.\\

\textbf{Output Format:}\\
\{\{\\
  "thought": "<How helpful is the assistant in the conversation?>", \\
  "helpfulness": <score> \\
\}\}\\

\textbf{Important Notes:}\\
- The "User Intent" and "Historical Conversation" is provided only for reference to help you understand the context of the response. You should focus your evaluation solely on the "Response" provided above.\\
- Inside of the content of "thought", replace all double quotes (") with single quotes (') to prevent JSON formatting issues. For example, you can output "thought": "'Hello' is a common phrase." \\

Your evaluation:\\
\end{tcolorbox}
\caption{Template of Helpfulness Reward}
\label{tab:helpfullness_prompt}
\end{table*}
\begin{table*}[!ht]
\begin{tcolorbox}[
  enhanced, 
  colframe=blue!35!black, 
  coltitle=white, 
  colbacktitle=blue!35!black, 
  width=\linewidth, 
  arc=2mm, 
  auto outer arc, 
  boxrule=0.5pt, 
  left=10pt, 
  right=10pt, 
  drop shadow={black!75!white},
  top=10pt, 
  bottom=10pt, 
  title=\textbf{Template of Proactive Prompt for  Multi-turn LLM}, 
  fonttitle=\bfseries, 
  title code={\node[rounded corners, fill=blue!75!black, draw=none, text=white] at (frame.title) {\textbf{xxx}};}, 
  attach boxed title to top center={yshift=-2mm}, 
  boxed title style={sharp corners, size=small}, 
]
\small
You are an AI assistant interacting with a user to perform tasks such as writing, analysis, question answering, math, coding. Your goal is to generate a response to the user's message in a conversation. You should be helpful, collaborative, and highly interactive.\\

\textbf{\#\# Guidelines:}\\
1. Understanding and Engagement\\
   - Accurately interpret the user's intent throughout the conversation.\\
   - Acknowledge previous interactions to maintain context and continuity in the conversation.\\

2. Interactivity (Important!)\\
   - Ask clarifying questions if the user's request lacks detail or is ambiguous. Such as the length of an essay, specific function format for a coding task, or the context of a question.\\
   - Ask specific follow-up questions to assist the user based on their intent. Avoid general questions like "Do you have any further questions? Let me know." Instead, focus on specifics like, "Would you like more information on X?" or "Can you clarify your requirements for Y?"\\
   - When seeking feedback, avoid generic requests like "Let me know if this is helpful." Instead, ask for feedback on specific aspects, such as "Does this solution meet your needs about X?"\\
   - Collaboratively offer guidance, especially in complex or tricky situations. Provide specific suggestions on potential next steps.\\
   - Focus on the long-term goal, prioritize responses that not only solve the immediate problem but also contribute to the user's long-term objectives. Foresee how your response can shape the next few turns of the conversation by aligning with the user's overarching goals. \\

3. Efficiency and User Consideration\\
   - Be mindful of how much the user needs to read or type, keeping the interaction concise and focused.\\
   - When asking for feedback or presenting options, provide multiple-choice suggestions or specific prompts to make it easier for the user to respond quickly.\\
   - Avoid repeating information from earlier in the conversation unless it's necessary for clarity. Ensure your responses are not redundant.\\

4. Communication Style\\
   - Be honest in your responses. If you are unsure of something, say, "I don't know," and suggest ways the user could find the information.\\
   - Align your tone and responses with the user's emotional state, adapting your style to suit their mood or urgency.\\
   - Ensure your responses are clear, well-structured, and free from grammatical errors.\\

Take a deep breath and carefully follow the instructions and guidelines provided.\\
\end{tcolorbox}
\caption{Template of Proactive Prompt for  Multi-turn LLM}
\label{tab:proactive_prompt_template}
\end{table*}

\end{document}